\documentclass[a4paper,fleqn]{cas-dc}
\usepackage[square,numbers]{natbib}
\usepackage{balance}
\usepackage{color,soul}
\usepackage{graphicx}
\usepackage{amstext}
\usepackage{amsmath}
\usepackage{amssymb}
\usepackage{booktabs}
\usepackage{caption}
\usepackage{subcaption}
\usepackage{hyperref}
\usepackage{url}
\usepackage{multirow}
\usepackage{multicol}
\usepackage{algorithm2e}
\usepackage{academicons}
\usepackage{xcolor}
\usepackage{xspace}

\def\tsc#1{\csdef{#1}{\textsc{\lowercase{#1}}\xspace}}
\tsc{WGM}
\tsc{QE}
\tsc{EP}
\tsc{PMS}
\tsc{BEC}
\tsc{DE}

\begin{document}

\let\WriteBookmarks\relax
\def\floatpagepagefraction{1}
\def\textpagefraction{.001}

\shorttitle{Involution Fused ConvNet for Classifying Eye-Tracking Patterns of Children with ASD}


\title [mode = title]{Involution Fused ConvNet for Classifying Eye-Tracking Patterns of Children with Autism Spectrum Disorder}                      
\tnotemark[]



\shortauthors{Islam et al.}

                   
\author[1]{Md. Farhadul Islam}[orcid=0000-0003-3249-4490]
\cormark[1]
\ead{farhadul.islam@bracu.ac.bd}

\author[1, 4]{Meem Arafat Manab}[orcid=0000-0002-2336-4160]
\fnmark[1]
\ead{meem.arafat@bracu.ac.bd}

\author[1, 5]{Joyanta Jyoti Mondal}[orcid=0000-0003-3113-8603]
\fnmark[1]
\ead{jmondal@uab.edu}

\author[1, 3]{Sarah Zabeen}[orcid=0000-0002-9406-3816]
\ead{sarah.zabeen@bracu.ac.bd}

\author[2]{Fardin Bin Rahman}[orcid=0000-0002-3771-7000]
\ead{fardin.bin.rahman@g.bracu.ac.bd}

\author[6]{Md Zahidul Hasan}
\ead{zadid.hasan@bracu.ac.bd}

\author[2]{Farig Sadeque}[orcid=0000-0001-6797-7826]
\ead{farig.sadeque@bracu.ac.bd}

\author[1]{Jannatun Noor}[orcid=0000-0001-9669-151X]
\ead{jannatun.noor@bracu.ac.bd}

\cortext[cor1]{Corresponding Author}
\fntext[fn1]{Equal Contribution}

\address[1]{Computing for Sustainability and Social Good (C2SG) Research Group, Department of Computer Science and Engineering, School of Data and Sciences, BRAC University, Dhaka, Bangladesh}
\address[2]{Department of Computer Science and Engineering, School of Data and Sciences, BRAC University, Dhaka, Bangladesh}
\address[3]{Department of Mathematics and Natural Sciences, School of Data and Sciences, BRAC University, Dhaka, Bangladesh}
\address[4]{School of Law and Government, Dublin City University, Dublin, Ireland}
\address[5]{Department of Computer Science, College of Arts and Sciences, University of Alabama at Birmingham, United
States}
\address[6]{Concordia Institute for Information Systems Engineering (CIISE), Gina Cody School of Engineering and Computer Science, Concordia University, Canada}

\begin{abstract}
Autism Spectrum Disorder (ASD) is a complicated neurological condition which is challenging to diagnose. Numerous studies demonstrate that children diagnosed with autism struggle with maintaining attention spans and have less focused vision. The eye-tracking technology has drawn special attention in the context of ASD since anomalies in gaze have long been acknowledged as a defining feature of autism in general. Deep Learning (DL) approaches coupled with eye-tracking sensors are exploiting additional capabilities to advance the diagnostic and its applications. By learning intricate nonlinear input-output relations, DL can accurately recognize the various gaze and eye-tracking patterns and adjust to the data. Convolutions alone are insufficient to capture the important spatial information in gaze patterns or eye tracking. The dynamic kernel-based process known as involutions can improve the efficiency of classifying gaze patterns or eye tracking data. In this paper, we utilise two different image-processing operations to see how these processes learn eye-tracking patterns. Since these patterns are primarily based on spatial information, we use involution with convolution making it a hybrid, which adds location-specific capability to a deep learning model. Our proposed model is implemented in a simple yet effective approach, which makes it easier for applying in real life. We investigate the reasons why our approach works well for classifying eye-tracking patterns. For comparative analysis, we experiment with two separate datasets as well as a combined version of both. The results show that IC with three involution layers outperforms the previous approaches.

\end{abstract}




\maketitle
\section{Introduction}

Autism Spectrum Disorder ASD is a term denoted to describe a group of complex neurodevelopmental conditions such as impairments in social communication, deficits in social interactions, and restricted, repetitive behavioural patterns \cite{wing1979severe}. 
Individuals with ASD exhibit a variety of symptoms as there are no specific biomarkers for this disorder. Commonly, those diagnosed with ASD experience difficulties in social communication and often struggle to maintain focus on certain tasks \cite{christensen2018prevalence}. These challenging symptoms can significantly impact their social and personal development, adversely affecting their daily lives. According to recent studies, ASD now affects 1 in every 50 to 68 children \cite{christensen2018prevalence,new6}. Changes in diagnostic criteria have been a major factor in this growth, but other factors include societal efforts to raise awareness and educate society. Mandy et al. \cite{asd_new} demonstrate how social issues are exacerbated throughout the lives of children with late-diagnosed autism. Prior to receiving an autism diagnosis, late-diagnosed autistic children frequently struggled greatly with their mental health and social skills, and they frequently got worse as they approached puberty.

Eye-tracking is the process of capturing, tracking and measuring eye movements or the absolute point of gaze (POG), which refers to the point where the eye gaze is focused in the visual scene \cite{majaranta_eye_2014}. The eye-tracking technology received particular attention in the ASD context since abnormalities of gaze have been consistently recognized as the hallmark of autism in general. The Psychology literature is replete with studies that analyzed eye movements in response to verbal or visual cues as signs of ASD \cite{coonrod2004early,jones2014developmental,sepeta2012abnormal,kylliainen2012affective}.

Studies on eye movement, specifically saccades and fixations, have proven valuable in identifying various mental states, cognitive functions, and neurological disorders, including ASD \cite{a7,a18}. Eye movement patterns in individuals with autism can help in recognizing ASD characteristics and assist in diagnosis. Existing research in this area has mostly focused on limited types of stimuli, such as faces, and there is a scarcity of publicly available datasets on eye movements in children with ASD. Public datasets are vital for advancing research in visual attention \cite{a4,a9,a12}, particularly for developing models tailored to the gaze patterns of individuals with ASD and for improving ASD diagnosis. Models specific to visual attention in ASD are crucial for understanding ASD traits, enhancing overall comprehension of ASD, and creating specialized resources, like educational materials, for those with ASD \cite{a19}. Additionally, these models can differentiate between individuals with ASD and healthy controls using gaze data, offering an alternative to the current expensive, subjective, and time-consuming diagnostic methods for ASD. The use of visual attention techniques could significantly enhance the efficiency of ASD diagnosis.

The combination of eye-tracking equipment and contemporary AI algorithms has improved diagnostic and application capabilities. When seeking an additional perspective that is thought to be more impartial and consistent, data-driven methods like Machine Learning (ML) are being accepted with greater vigor. Nowadays, evaluations involve handling non-traditional data that can be gathered from observations originating via electrophysiology or chemical laboratory experiments. These technologies include, for instance, visual tracking imaging equipment and the use of virtual reality. Researchers have studied the brain networks as well as sentiment identification in order to determine the functional connections and distinguish between individuals having ASD and those who are believed to be developing normally \cite{Shirwaikar_2023, UDDIN2024107185}. In order to help with the identification of ASD, this endeavor continues along the route of combining ML with eye-tracking technologies. Research divisions in AI and psychology are working as an interdisciplinary collaboration on this project. Our method is unique in that it relies on the concept that visual depictions of gaze tracking scanpaths may distinguish autism-related gaze behavior.

High consistency between data patterns is provided by the ML algorithm for the categorization of ASD. Through the linkage of multiple regions, a structurally linked area of the human brain was discovered. Techniques based on deep learning have been presented in recent years as a result of automated feature learning and processing complicated data \cite{f18,f19,20}. For greater accuracy in classification, unsupervised training of layered auto encoders are employed in deep learning approaches like Multi-Layer Perceptrons (MLP) \cite{21,22}. In order to calculate the high level connection among these characteristics, ML models employ several voxels as input \cite{23}. SVM, Naive Bays, K-nearest-neighbor algorithm (KNN), CNN, and logistical regression are among the ML algorithms used to forecast and evaluate ASD. To improve the accuracy of classification while using the least amount of feature subsets possible, artificial intelligence algorithms are used alongside the machine learning model \cite{24}.

Using state-of-the-art ML models also has issues, considering most of the models are resource heavy. The complexity of a deep model is affected by its size. The number of parameters, hidden layers, hidden layer width, number of filters, and filter size are a few often used metrics for measuring model size. Models of varying sizes can have their complexities measured using the same complexity measure criteria within the same model framework, making them comparable \cite{Kůrková2018}.

Inference can be computationally demanding in certain works due to the usage of progressively bigger models, for instance, with respect to model parameters. This makes training these models even more demanding. Similar to run time, there is a correlation between this metric and work. It is independent of the underlying hardware. Furthermore, there is a strong correlation seen here between this metric and the model's memory usage. However, various algorithms employ their parameters differently. For example, by deepening versus expanding the model. Because of this, various models with comparable numbers of parameters frequently carry out varying amounts of work. The emphasis on this one statistic overlooks the social, environmental, and economic costs associated with utilizing heavyweight models to achieve the claimed outcomes, even in spite of the obvious benefits of increasing model accuracy \cite{3381831}. In this case, efficient lightweight models are significant for a greener environment. According to Menghani \cite{3578938} there are some challenges faced by a DL practitioner when training or deploying a model, which are:
\begin{itemize}
    \item Sustainable server-side scaling: Large-scale deep learning model deployment and training are an expensive set of tasks. While fees for this kind of training may be one-off, (or free if the model is already trained), the cost of installing and allowing inference to operate for an extended amount of time might still be high due to the usage of RAM, CPU, and other resources on the server. Even companies such as Amazon, Google, Facebook, etc, which have annual expenditures on capital that exceed billions of dollars, have legitimate concerns over the environmental impact of their data centers. 
    \item Enabling on-device deployment: In order to accommodate a variety of considerations (confidentiality, connection, sensitivity), certain applications that use deep learning must operate in real-time mode in compliance with IoT and smart devices. Here, the model inference takes place directly on the gadget. It is therefore crucial to adapt and customize models for the intended gadgets. 
    \item Privacy and data sensitivity: When user data is particularly sensitive, the ability to utilize as minimal information as feasible for training is crucial. Therefore, if models can be trained well with a reduced amount of data, then a lesser amount of information gathering is necessary.  
    \item New applications: Some novel applications possess new restrictions (pertaining to quality of model or footprint) which current commercial models may not be able to meet.
\end{itemize}

Therefore, making a suitable model for efficiently extracting gaze patterns will require a special architectural combination that will focus on these particular type of data while keeping the model size as small as possible.

Our problem focuses on eye-tracking data. Our assumption is that eye-tracking produces highly localized data (cf. figure 2) where most paths seem to converge in a specific direction of the images. Previous research mostly utilized CNNs to classify eye-tracking saliency map data for ASD detection \cite{111}, but as the original eye-tracking data is more localized, scattered, and globally variable, and as involution has proven to be more effective than convolution architecture in preserving spatial contrasts and spread-out information \cite{112}, we believed that involution layers would increase the performance of the DL architecture. Previously, involution neural networks proved very effective for similarly structured data from widely different sources, such as point clouds, skin lesions and malaria paratisized cells \cite{112, 113, unic}. Unlike convolution kernels, which represent channel-specific and location-agnostic features in the data images, involution kernels can identify location-specific and channel-agnostic features in the image set, which can be of particular help in eye-tracking data, as our analysis later shows. 

Because of its linear complexity, involution more than makes up for not modelling inter-pixel interactions the way attention does \cite{inn_main}. The weights of several kernels no longer need to be maintained. Finding the meta weights is all that is required. This makes involution contain very less amount of weight parameters, almost negligible comparing to convolution. Compared to convolution, this enables the construction of bigger models. Moreover, boosting convolution-based models' performance by adding involution results adding very few weight parameters. This does not make the model bigger in terms of storage but may boost the performance.

There have been several works related to classifying ASD \cite{akter2021improved,raj2020analysis,xie2019two,biswas_xai_2021}. To the best of our knowledge and from our literature study, we see that this is the first work that involves the Involution process to classify ASD with eye-tracking or saliency maps. There have been a few works regarding the combination of Involution and Convolution \cite{icnet,ci,unic}. None of them focus on eye-tracking images, moreover the construction of the hybrid model differs, and the design has different motives as well. Islam et al. \cite{unic} propose a hybrid involution-convolution architecture that only uses a single layer of involution. The study focuses on cell-like medical images, which have an anomaly that indicates the region of disease. To classify these two, a location-specific operation helps the process effectively. But overuse of the location-specific process of involution leads to overfitting. 

In this study, the data requires a heavier approach to the location-specific operation. This preference for involution architecture does not preclude the need for convolution layers. The convolution kernels provide us with an effective way of extracting a feature map that is shift-invariant, leading us to detect shapes better, while involution helps us in location-dependent anomaly detection. But involution alone can make the model comparatively larger after flattening the inputs for Dense layers. Having convolution operations with maxpooling will decrease the weights in a significant way for both performance and size. Convolution will be helpful to recover smoothness in the approximation as well as reduces the error \cite{203}. Thus, a combination of involution and convolution networks would be the most effective in ASD detection from eye-tracking images, which have both translation-invariant and anomalous features.

People with ASD exhibit distinctive patterns in their gaze, which can be detected through the visual analysis of eye-tracking data \cite{falck2013eye}. So, in this study, our key contributions include how impactful involutions can be in the case of learning eye-tracking patterns. We propose an Involution-Convolution architecture for this task. We show how both processes result in good performance. For experimental purposes, we utilise two different datasets and merge them into one, which shows how our proposed model boosts performance. Both of the datasets are equally balanced, and the ASD class images have a similar pattern. Since ASD diagnosed children have less focus on looking at a particular object, their focus is distributed over the span of their eyesight. We show how involutions effectively learn this pattern with less impact on the size of the model. Moreover, we show how the number of involution layers affects the classification performance up to a limit.

The contribution and novelty of our study are as follows:

\begin{itemize}
    \item We propose an easy-to-build and novel hybrid model based on involution and convolution. Having both location specific and channel specific operations and also comparatively lightweight (in terms of weight parameters).

    \item Our model is storage efficient, requiring only 1.36 MB. Even though convolution is a heavier approach,  our model is designed in such a way that it remains smaller than involution-only model.
    
    \item We analyze the data to show why location specific approach is effective in eye-tracking data.

    \item We also show how many involution layers are required for this study in both qualitative and quantitative ways.

    \item Our model outperforms popular models and previous works as well having an accuracy of 99.43\% in one dataset \cite{Elbattah2019-ri} and 96.78\% \cite{Duan2019-tl} in another. We also show how our model performs well on different types of data tested together, showing how our model may perform well on multimodal data.
\end{itemize}






The remainder of the paper is divided into the following sections: The literature review is in Section~\ref{Literature Review}. Then, in Section~\ref{Research Methodology}, we provide a rundown of our utilized dataset, data analysis, proposed classification architecture. We describe our experimental setup, results, ablation studies and comparison in Section~\ref{Experiments}.
Subsequently, we go to our research findings, significance and limitations in Section~\ref{Discussion}. Finally, we explore potential future works and conclude the paper in Section~\ref{Conclusion and Future Work}. 

\section{Literature Review}
\label{Literature Review}

Given the serious nature of the issue, numerous studies on recognizing signs of ASD have been carried out throughout the years. Nevertheless, several ASD related behavioral and social symptoms, such as sociability, diminished attention, eye contact, and recognition of facial features, might be difficult to detect in the early stages. As a result, even established clinical procedures are likely to have limits in terms of an early diagnosis, which could delay the rehabilitation process. Given the varied degrees of intensity of the symptoms, identifying the issue is highly difficult and requires additional examination. Previous studies mostly use Convolutional Neural Networks (CNN) as their most dominant methodology.

Akter et al. \cite{akter2021improved} devise an approach that recognizes ASD by utilising transfer learning in the case of examining facial characteristics. They designed an enhanced system for facial recognition by employing transfer learning, which is capable of identifying people with ASD. Several deep learning and machine learning models were used, with the upgraded MobileNet-V1 model achieving the highest accuracy of 90.67\%. Additionally, they employed the k-means clustering technique, altering the value of k from 2 to 10, to identify the various ASD groups. Subsequently, the proposed model is employed to forecast the sub-types, resulting in an accuracy rate of 92.10\% under the condition of k = 2. While the suggested model demonstrated a commendable level of accuracy, it is important to acknowledge that face recognition systems may encounter limitations, particularly during the first stages of development. 

Raj et al. \cite{raj2020analysis} employ several machine learning and deep learning methodologies, including Naive Bayes, Support Vector Machine, Logistic Regression, K-Nearest Neighbours, Artificial Neural Network (ANN), and CNN, for the purpose of detecting Autism Spectrum Disorder (ASD) in children. Authors use three publicly accessible datasets sourced from the UCI Repository. Each dataset referred to distinct age cohorts. The dataset utilised in this study comprises a total of twenty variables, predominantly encompassing demographic and screening-related data. Based on the findings of their experiment, it was seen that the CNN model exhibited superior performance compared to the other models. The CNN model achieved accuracy rates of 99.53\%, 98.30\%, and 96.88\% for the adult, children, and adolescents datasets, respectively.

Xie et al. \cite{xie2019two} develop a two-stream deep learning network to detect visual attention in ASD. Two identical VGGNets that were modified from the VGG16 architecture served as the basis for the proposed framework. Data for this study was gathered from the OSIE database, which has 700 photographs of people with ASD and TD's eye movement patterns. The proposed model's accuracy was 95\% and 85\%, respectively. Additionally, they used t-SNE visualisations to find a few pixel-level characteristics in the images.


Ahmed et al. \cite{Ahmed2020-rw} investigate the use of eye-tracking technology to detect ASD by analyzing atypical visual attention in children. They propose a deep learning model based on a CNN that can classify children as autistic or typically-developing based on their eye-tracking scanpaths. Tested on a sample of 29 autistic and 30 typically-developing children, the model demonstrated a high testing accuracy of 98\%, indicating its potential effectiveness in the accurate diagnosis of ASD through the observation of visual attention patterns. Their extended work \cite{Ahmed2023-wu} focuses on the critical role of early detection and diagnosis in improving developmental outcomes for individuals with ASD. It explores the integration of eye-tracking technology and deep learning algorithms to distinguish between children with ASD and their typically developing peers. The study introduces deep learning models based on CNN and Recurrent Neural Network (RNN) architectures, utilizing an eye-tracking dataset for training. The results are promising, with the Bidirectional Long Short-Term Memory (BiLSTM), Gated Recurrent Unit (GRU), CNN-LSTM hybrid, and LSTM models achieving high accuracy rates. 

\begin{figure*}[]
\centering
\begin{subfigure}{.45\textwidth}
    \centering
    \includegraphics[width=.9\linewidth]{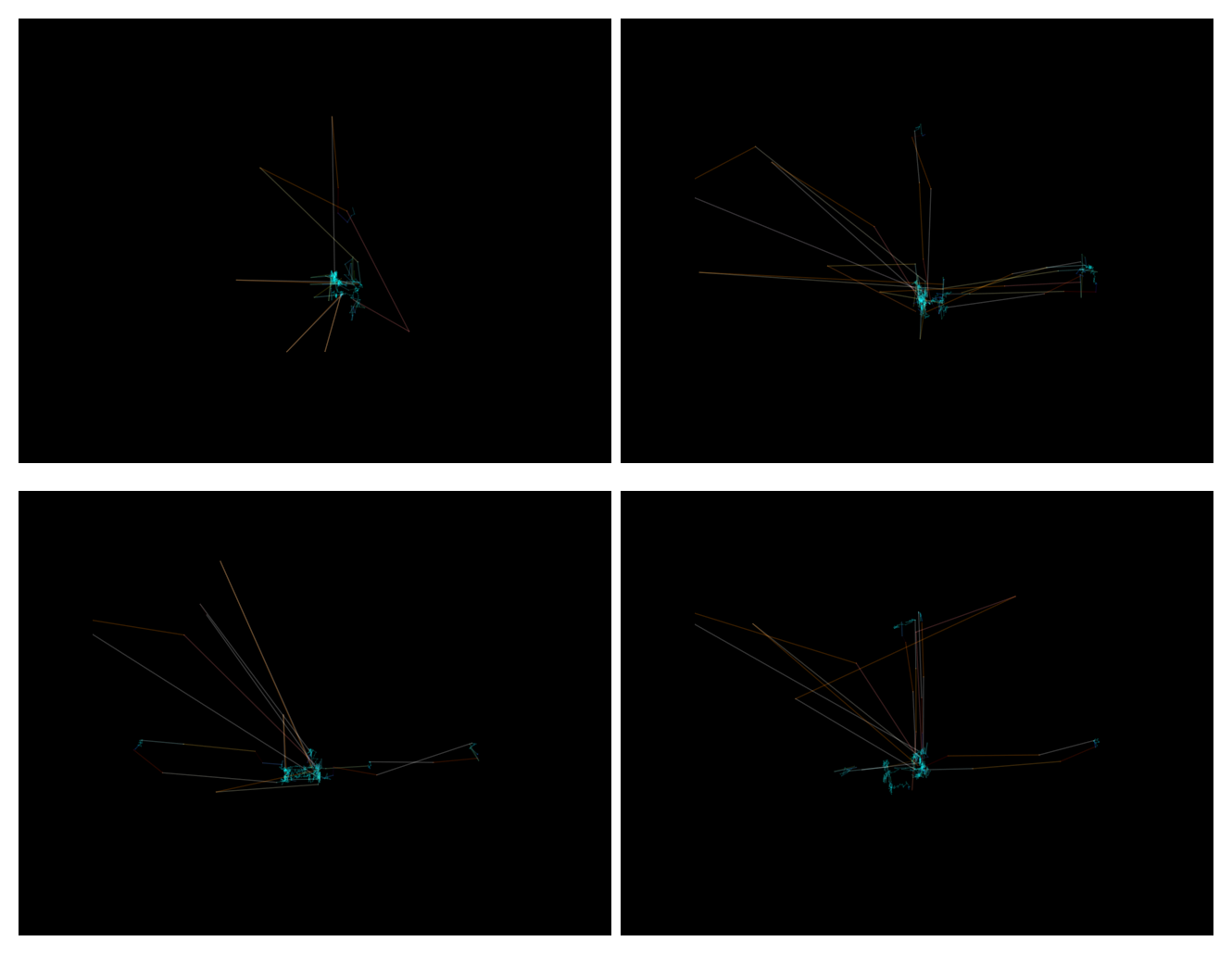}
    \caption{Samples images from the TD class of dataset-1}
    \label{sample_data-1}
\end{subfigure}
\begin{subfigure}{.45\textwidth}
    \centering
    \includegraphics[width=.9\linewidth]{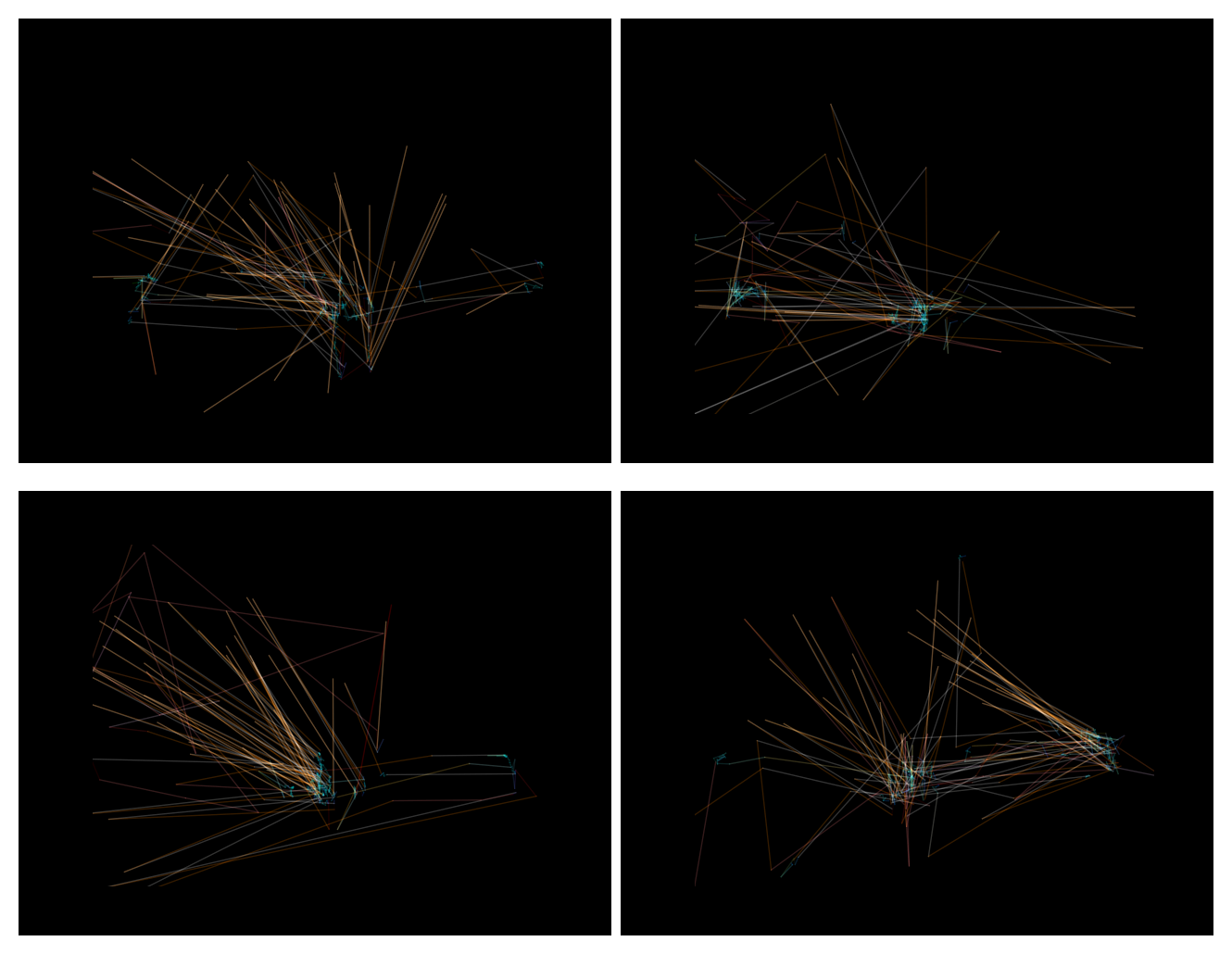}
    \caption{Samples images from the ASD class of dataset-1}
    \label{sample_data-2}
\end{subfigure}
\begin{subfigure}{.45\textwidth}
    \centering
    \includegraphics[width=.9\linewidth]{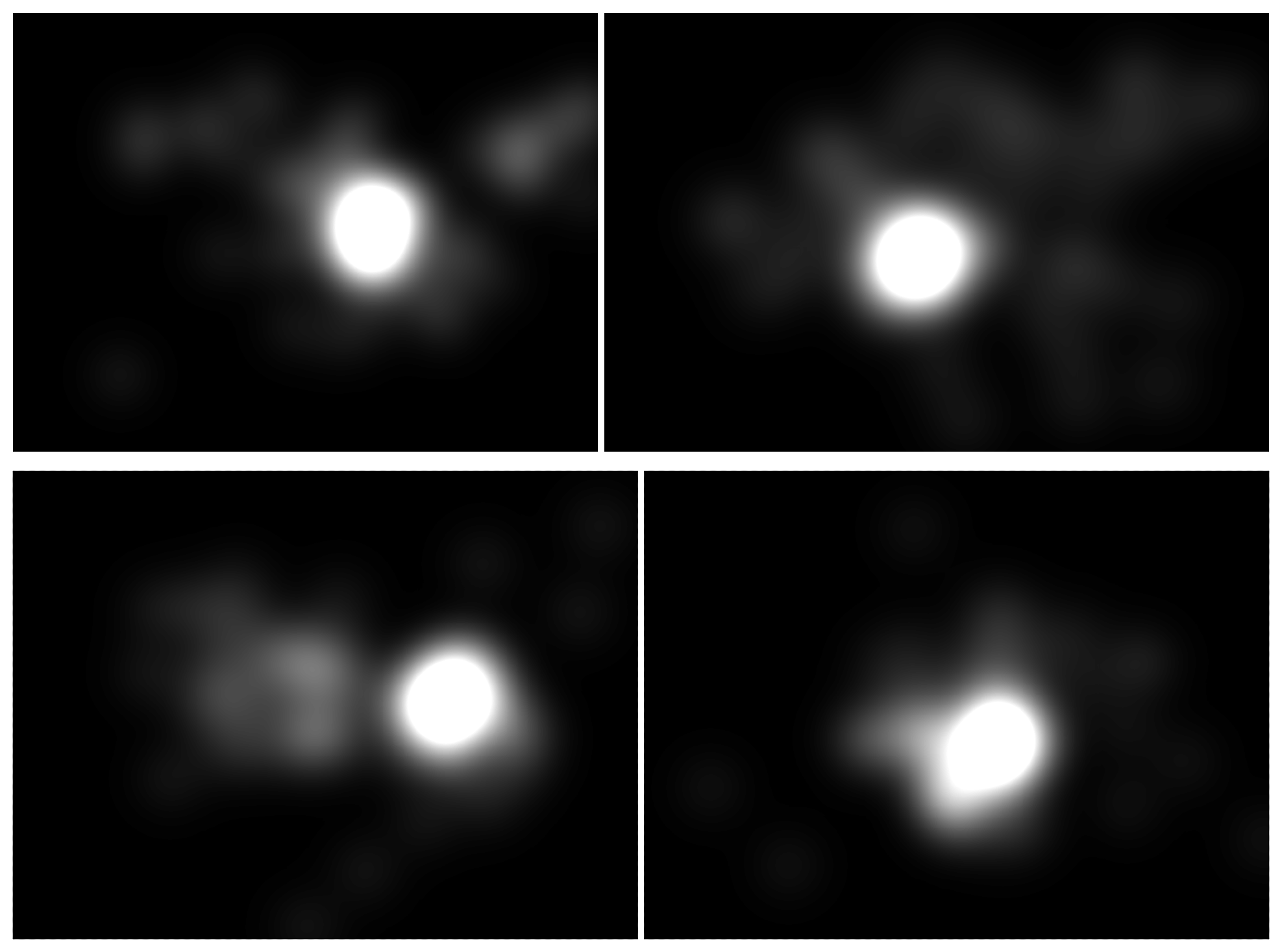}
    \caption{Samples images from the TD class of dataset-2}
    \label{sample_data-3}
\end{subfigure}
\begin{subfigure}{.45\textwidth}
    \centering
    \includegraphics[width=.9\linewidth]{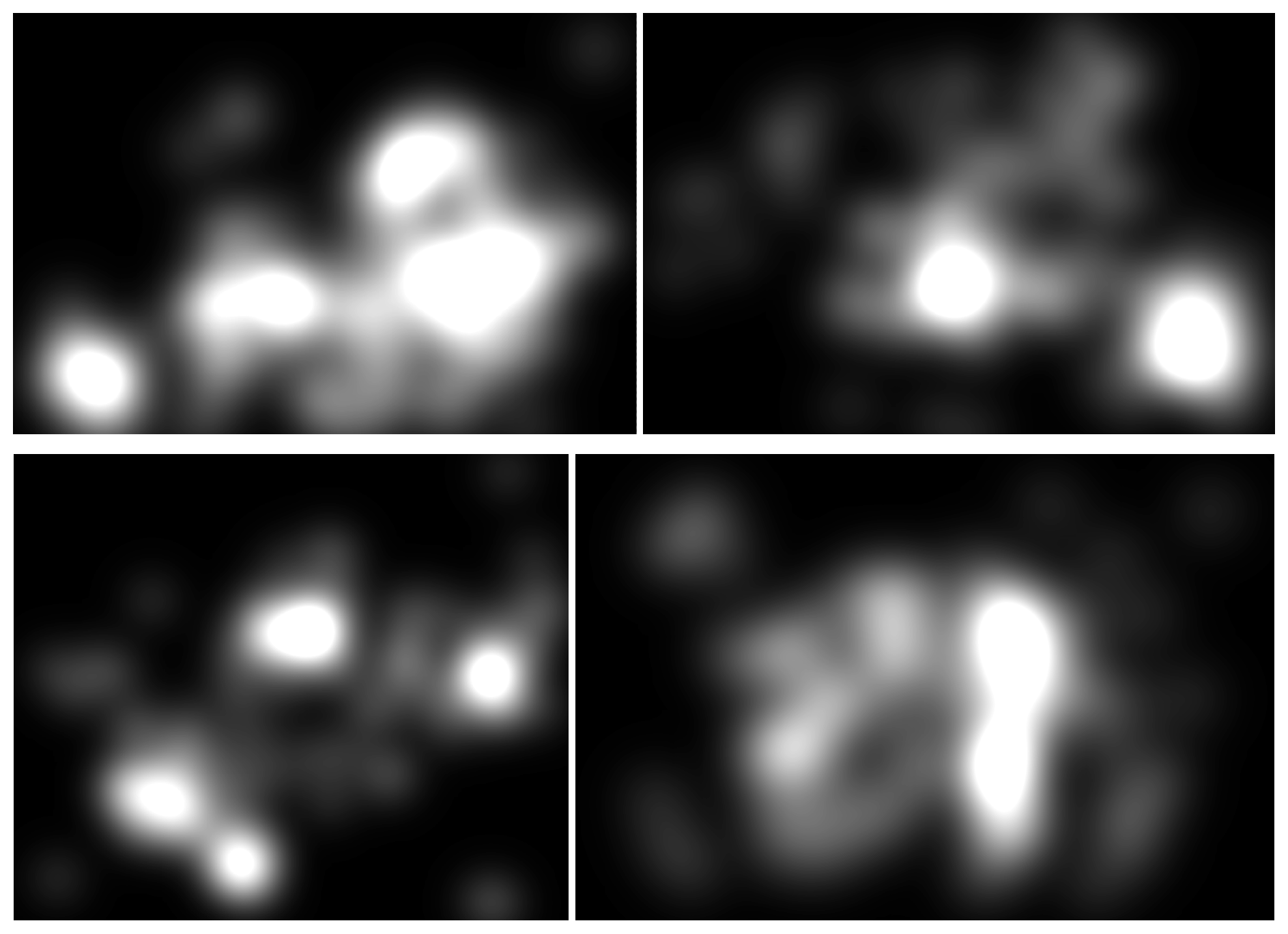}
    \caption{Samples images from the ASD class of dataset-2}
    \label{sample_data-4}
\end{subfigure}
\caption{Samples images from dataset-1 and dataset-2.}
\label{sample_data}
\end{figure*}

\begin{figure*}
\centering
\begin{subfigure}{.5\textwidth}
  \centering
  \includegraphics [width=8.7cm, height=6cm]{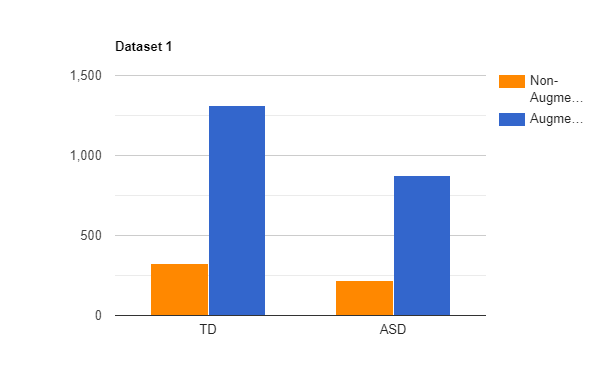}
  \caption{}
  \label{fig_2-1}
\end{subfigure}%
\begin{subfigure}{.5\textwidth}
  \centering
  \includegraphics [width=8.7cm, height=6cm]{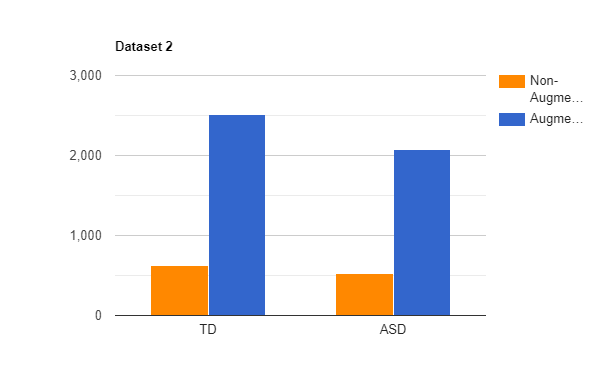}
  \caption{}
  \label{fig_2-2}
\end{subfigure}
\caption{Bar chart of data frequency of both augmented and non-augmented versions.}
\label{fig_2}
\end{figure*}

Jain et al. \cite{10287306} present an approach for the early diagnosis of ASD using Magnetic Resonance Imaging (MRI) brain images. The study introduces a Deep CNN enhanced with a Dwarf Mongoose optimized Residual Network (DM-ResNet) for classifying ASD. The process begins with preprocessing the MRI brain images to remove non-brain tissues, followed by segmentation using a hybrid of Fuzzy C Means (FCM) and Gaussian Mixture Model (GMM). This segmentation partitions the image into cortical and subcortical regions, simplifying the classification task. VGG-16 networks are then used for extracting intricate and discriminative features from these segmented images. Additionally, a Region of Interest (ROI) based functional connectivity feature extraction is conducted with VGG-16. These features are classified using the DM optimized ResNet, with hyperparameters fine-tuned through the DM optimization algorithm to enhance classification accuracy. The proposed method significantly improves the accuracy of autism detection, reaching a remarkable 99.83\%.

Ahmed et al. \cite{electronics11040530} employs neural networks (specifically FeedForward Neural Networks (FFNNs) and ANNs, utilizing a combination of Local Binary Pattern (LBP) and Grey Level Co-occurrence Matrix (GLCM) algorithms for feature classification. This method achieved an impressive 99.8\% accuracy. The second technique involves pre-trained CNN models like GoogleNet and ResNet-18, focusing on deep feature map extraction, with GoogleNet and ResNet-18 models attaining high accuracies of 93.6\% and 97.6\% respectively. The third technique, a hybrid method, combines DL [GoogleNet and ResNet-18] with machine learning [Support Vector Machine (SVM)]. This approach uses CNN for deep feature map extraction and SVM for feature classification, resulting in accuracies of 95.5\% for GoogleNet + SVM and 94.5\% for ResNet-18 + SVM, demonstrating the potential of these AI techniques in accurately diagnosing ASD through analysis of visual behavior.

Elbattah et al. \cite{Elbattah2022-xz} investigate the application of Transfer Learning (TL) in the detection of ASD, an innovative approach combining TL with eye-tracking technology. The study tests various established models, including VGG-16, ResNet, and DenseNet, to evaluate this methodology. The results demonstrate a promising accuracy in classification, with the ROC-AUC reaching up to 0.78. While the study does not claim to outperform existing methods, it highlights an intriguing aspect of using (synthetic) visual representations of eye-tracking data. This approach leverages the power of models pretrained on extensive datasets like ImageNet.

To effectively classify eye-tracking pictures, Mumenin et al. \cite{Mumenin2023-pt} suggest an altered CNN model which could highlight noteworthy trends within the images. The investigation was carried out on both RGB (three channel) images and monochromatic (one channel) images. This significantly affects the dimension of the model's input layer parameters. The suggested model also surpasses the current models by achieving 97.41\% accuracy when applied to gray-scale photos. It is further observed that the F1 score of this model outshines other approaches with a 97.83\% score for gray-scale photos and a 97.12\% score for RGB photos.

\begin{figure*}[h]
\centering
\begin{subfigure}{.35\textwidth}
    \centering
    \includegraphics[width=.9\linewidth]{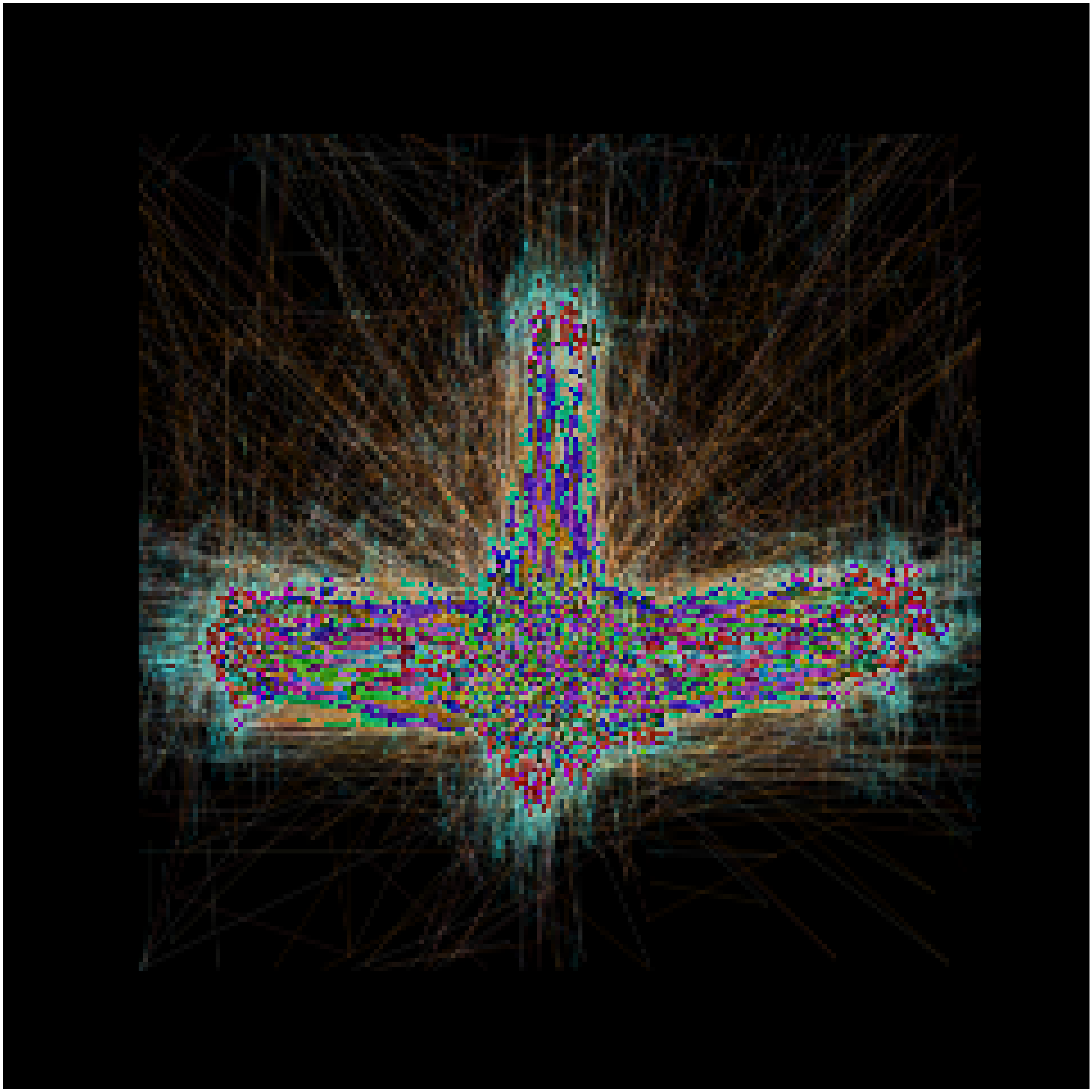}
    \caption{TD Mean Image of Dataset 1}
    \label{td_mean_1}
\end{subfigure}
\begin{subfigure}{.35\textwidth}
    \centering
    \includegraphics[width=.9\linewidth]{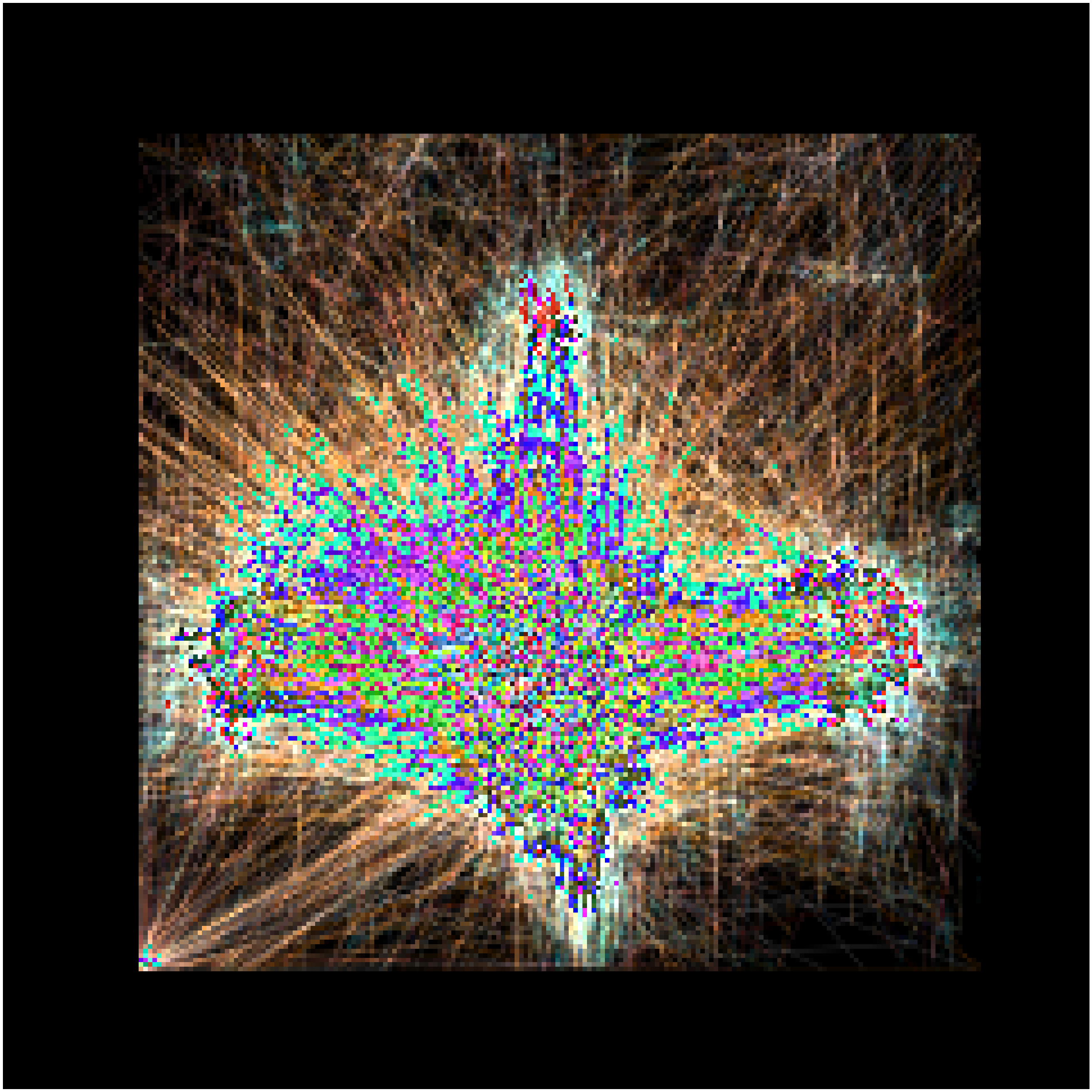}
    \caption{ASD Mean Image of Dataset 1}
    \label{asd_mean_1}
\end{subfigure}
\begin{subfigure}{.35\textwidth}
    \centering
    \includegraphics[width=.9\linewidth]{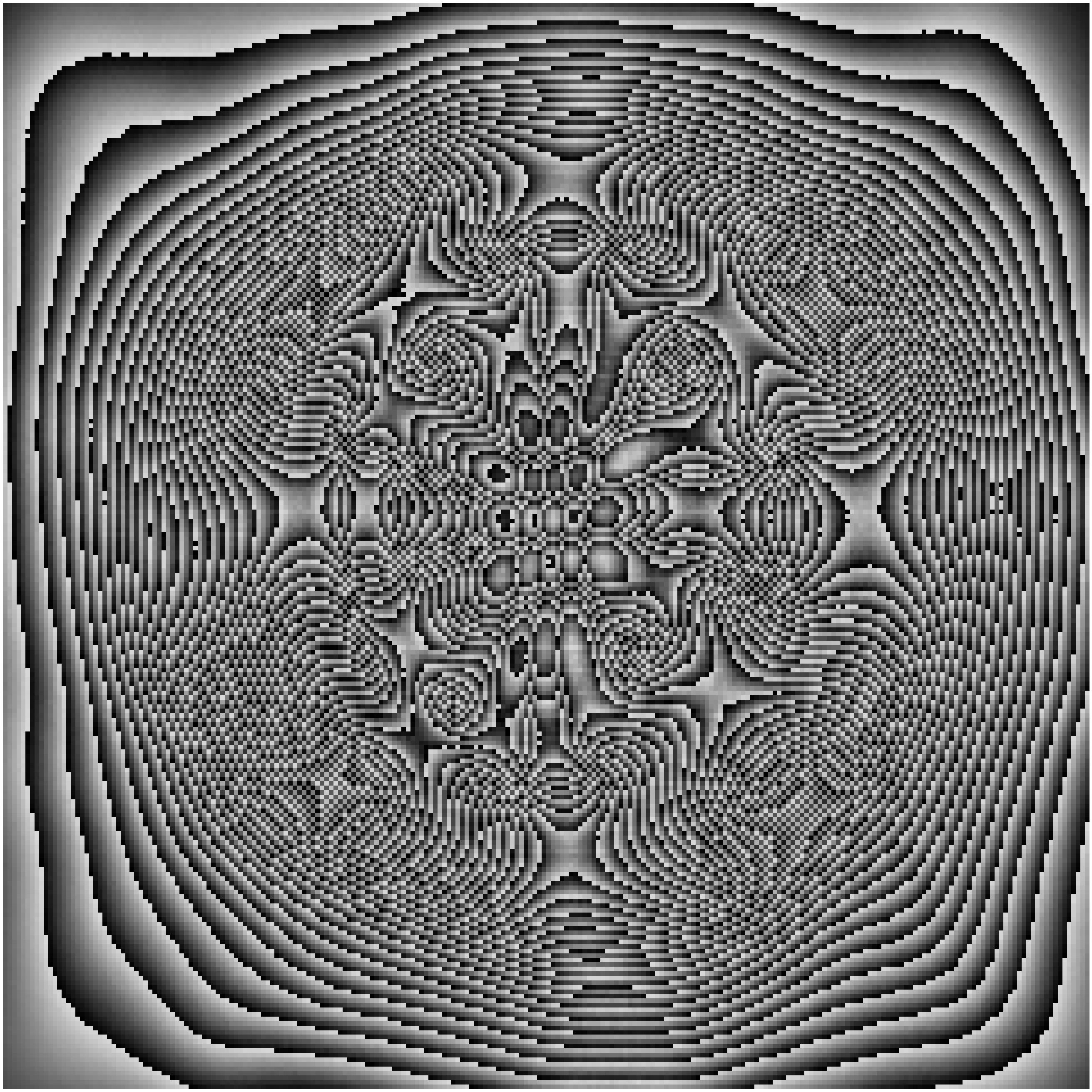}
    \caption{TD Mean Image of Dataset 2}
    \label{td_mean_2}
\end{subfigure}
\begin{subfigure}{.35\textwidth}
    \centering
    \includegraphics[width=.9\linewidth]{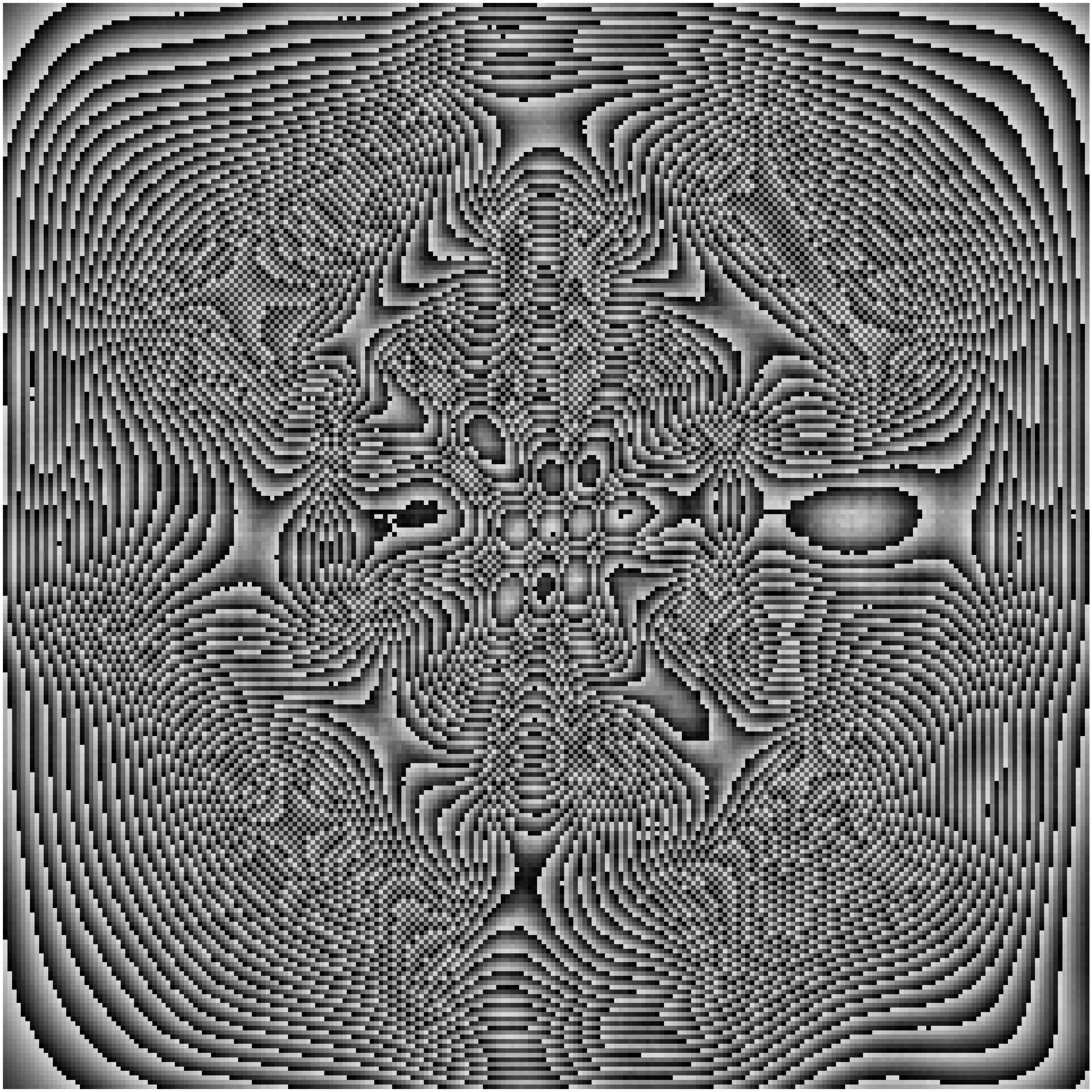}
    \caption{ASD Mean Image of Dataset 2}
    \label{asd_mean_2}
\end{subfigure}
\caption{Mean images of each class. These images were generated by finding the mean pixel value of each images per class.}
\label{data_sample}
\end{figure*}

From our literature study, we can conclude that the use of location-specific approaches is not common. Most of the studies emphasize the classification task without properly analyzing both the data and prediction characteristics. Moreover, these studies do not focus on model size and reliability in terms of architectural configuration. In our study, we focus on solving these issues.

\section{Research Methodology}
\label{Research Methodology}

\subsection{Dataset}

We use two datasets to experiment with our model. First dataset is created by Elbattah \cite{Elbattah2019-ri}. This dataset contains 547 images, where 328 images are from typically developed TD participants and 219 from ASD diagnosed participants. We refer to this dataset as, "Dataset 1". The second dataset is made by Duan et al. \cite{Duan2019-tl}. The dataset includes 300 images along with the eye movement data gathered from 14 children diagnosed with ASD and 14 children who are TD. We refer to this dataset as, "Dataset 2". To make the experiment more rigorous, we combine both datasets above and make a larger dataset, where 628 images are from TD participants and 519 are from ASD participants. In both cases, the frequency of each class is not very imbalanced. For experimental purposes and to check how the model learns, we augment the data using rotation, shear, and width shift techniques, which turn the dataset 4 times larger. 
Figure \ref{fig_2} illustrates the comparison between the amount of data in dataset-1 and dataset-2. The augmentation process was also done by the original authors of the dataset for experimental purposes. We also merge these two datasets to see if the models learn the spatial pattern of these two classes with different classes. The augmentation and merging process is not meant for diagnosis purposes. An increase in performance will indicate proper learning with location-specific processes. We split the dataset randomly into an 80:10:10 approach, where 80 is training and 10 for testing and validation data. The sample data are shown in Figure \ref{sample_data}. 


\begin{figure}[ht]
    \centering

        \includegraphics[width=0.7\linewidth]{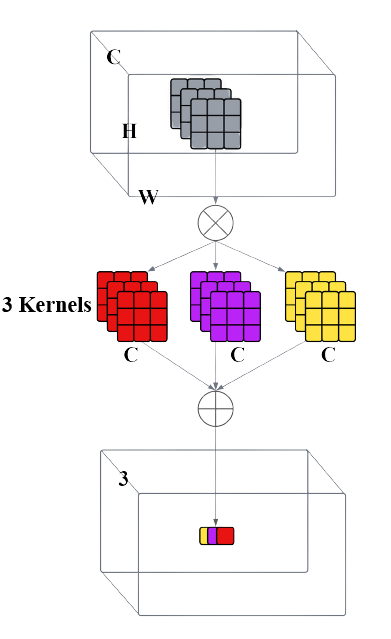}%
        \vspace{1cm}
   
        \caption{Visualization of the kernel production of convolution.}
    \label{conv}
        
\end{figure}

\begin{figure}[ht]
    \centering

        \includegraphics[width=0.8\linewidth]{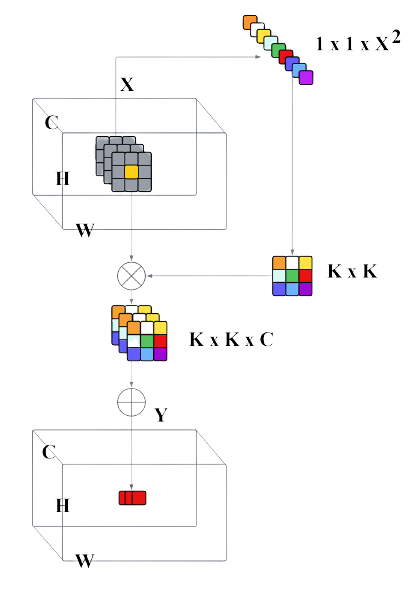}%
        \vspace{1cm}
   
        \caption{Visualization of the kernel production of involution.}
    \label{inv}
        
\end{figure}

\subsection{Data Analysis}

We analyze the data by finding the mean pixel values from all data and visualize to see the data characteristics. The visualizations can be seen in Figure \ref{data_sample}. When the mean images of the two classes (ASD), i.e. autism spectrum disorder and TD, i.e. typically developed) in the eye-tracking dataset were compared, a spatial bias was observed around the center of the images (see Figure \ref{td_mean_1} and \ref{asd_mean_1}) in dataset-1.  Moreover, the mean image for the ASD class was significantly more dispersed. Such highly visible results were not found in the saliency map dataset, but a pattern of centrality was still discernible (see Figure \ref{td_mean_2} and \ref{asd_mean_2}). The pattern from the first dataset in particular prompted us to use a location-specific layer in our deep-learning architecture, namely the involution layers. Data from both the classes (ASD and TD) for the first dataset show the dominant area of focus in both classes. The sight span of ASD is much broader than the other making the image more noisy. For the second dataset, a similar conclusion could be drawn, especially for ASD class where the image is much more nosiy than the other. This indicates that use of a location-specific approach \cite{involution_paper} may be useful to identify the different patterns of the ASD class.

\begin{figure*}[h]
    \makebox[\linewidth]{
        \includegraphics[width=\linewidth]{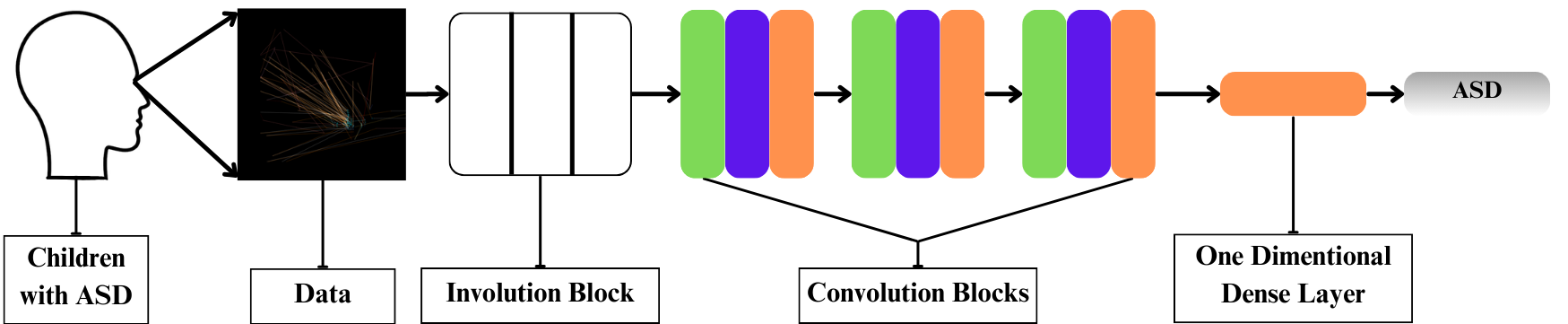}}

    \caption{Simplified illustration of the proposed architectural design. We simply put a single involution block before convolution based for extracting important spatial features beforehand.}
    \label{model}
\end{figure*}

\subsection{Classification Architecture}

Convolution kernels of size $K$, $K$, $C_{in}$, named $C_{out}$, and a tensor input, denoted by $X$ are used in the beginning of the process of convolution. $X$ has three dimensions $H$, $W$, and $C_{in}$. Through the addition and multiplication of the kernels with the input tensor, we obtain another tensor, $Y$ which has the dimensions $H$, $W$, and $C_{out}$. Figure \ref{conv} displays the value of $C_{out} =3 $. The dimensions $H$, $W$, and 3 for the output tensor are now available to us. Since the operation of the convolution kernel is not concerned with the location of the input tensor, it is regarded as location agnostic. Nevertheless, considering that each of the channels is built using a separate convolution filter, the output tensors that are produced are also unique.

\begin{table*}[]
\centering
\caption{Layer wise breakdown of weight parameters and output shape of the proposed model.}
\begin{tabular}{ccc}
\hline
\multicolumn{1}{c|}{\textbf{Layer (type)}} & \multicolumn{1}{c|}{Output Shape}                                      & Parameter \\ \hline
\multicolumn{1}{c|}{Input Layer}           & \multicolumn{1}{c|}{{[}(None, 48, 48, 3){]}}                           &       0    \\ 
\multicolumn{1}{c|}{Involution Layer}      & \multicolumn{1}{c|}{{[}(None, 48, 48, 3){]}, (None, 48, 48, 9, 1, 1))} &       26    \\ 
\multicolumn{1}{c|}{Involution Layer}      & \multicolumn{1}{c|}{{[}(None, 48, 48, 3){]}, (None, 48, 48, 9, 1, 1))} & 26        \\ 
\multicolumn{1}{c|}{Involution Layer}      & \multicolumn{1}{c|}{{[}(None, 48, 48, 3){]}, (None, 48, 48, 9, 1, 1))} & 26        \\

\multicolumn{1}{c|}{Convolution Layer}     & \multicolumn{1}{c|}{(None, 46, 46, 32)}                                & 896       \\ 
\multicolumn{1}{c|}{2D Max Pooling}        & \multicolumn{1}{c|}{(None, 23, 23, 32)}                                & 0         \\ 
\multicolumn{1}{c|}{Convolution Layer}     & \multicolumn{1}{c|}{(None, 21, 21, 64)}                                & 18496     \\ 
\multicolumn{1}{c|}{Batch Normalization}   & \multicolumn{1}{c|}{(None, 21, 21, 64)}                                & 256       \\ 
\multicolumn{1}{c|}{2D Max Pooling}        & \multicolumn{1}{c|}{(None, 10, 10, 64)}                                & 0         \\ 
\multicolumn{1}{c|}{Convolution Layer}     & \multicolumn{1}{c|}{(None, 8, 8, 128)}                                 & 73856     \\ 
\multicolumn{1}{c|}{Batch Normalization}   & \multicolumn{1}{c|}{(None, 8, 8, 128)}                                 & 512       \\ 
\multicolumn{1}{c|}{2D Max Pooling}        & \multicolumn{1}{c|}{(None, 4, 4, 128)}                                 & 0         \\ 
\multicolumn{1}{c|}{Flatten}               & \multicolumn{1}{c|}{(None, 2048)}                                      & 0         \\ 
\multicolumn{1}{c|}{Dense}                 & \multicolumn{1}{c|}{(None, 128)}                                       & 262272    \\ 
\multicolumn{1}{c|}{Dropout}               & \multicolumn{1}{c|}{(None, 128)}                                       & 0         \\ 
\multicolumn{1}{c|}{Dense}                 & \multicolumn{1}{c|}{(None, 2)}                                         & 258       \\  \hline
\multicolumn{3}{l}{Total parameters: 356,624 (1.36 MB)}        \\                                                                           
\multicolumn{3}{l}{Trainable parameters: 356,234 (1.36 MB)}      \\
\multicolumn{3}{l}{Non-trainable parameters: 390 (1.52 KB)}   \\ \hline
\end{tabular}
\label{tab_1}
\end{table*}

In Involution \cite{involution_paper}, the features of convolution are inverted. Except for the kernel, involution computation is identical to convolution computation. Unlike Convolution, a picture is not covered with a single learned kernel. Each pixel's kernel is created dynamically based on the pixel's value and learning parameters. This is useful because, similar to attention input, it regulates the kernel's weights. Involution calculates each output pixel by utilizing Equation \ref{eq1}:


\begin{equation}
\label{eq1}
Y_{i,k,k} = \sum_{(u,u)\in \Delta_{K} }^{} H_{i,j,u + \lfloor K/2 \rfloor , v + \lfloor K/2 \rfloor , \lceil kG/C \rceil } X_{i+u,j+v,k}
\end{equation}

In the channel dimension, groups of channels share the same kernel. The kernel is produced using the linear network shown below in In Equation \ref{eq2}.

\begin{equation}
\label{eq2}
 H_{i,j} = \phi(X_{i,j}) = W_{1} \sigma(W_{0}X_{i,j})
\end{equation}

The meta weights ($W_1$, $W_0$) of the kernel, i.e. weights used to construct the kernel, are shared across all the pixels, therefore shift-invariance of convolution is not fully lost. Involution does not describe interpixel interactions as effectively as attention does, but its linear complexity more than makes up for it. So, we no longer need to record the weights of so many kernels. Instead, we only need to know the meta weights. This enables us to construct more complex models than with convolution. As a result, the number of weight parameters is much less even if we use multiple layers of involution operations. But the training time slightly increases, due to the operation complexity. Figure \ref{inv} provides an intuition on this kernel generation method.

We use convolution, involution, and a combination of both for this study. The involution-convolution hybrid uses both location-specific and channel-specific capabilities which helps the model to learn the eye-tracking patterns effectively. Since, the location-specific features are dominant in these data, we use involution layers before the convolution layers and the consisting max-pooling operations in convolution blocks will result a reduction in weight parameters. 

Max pooling is a convolution method that creates a feature map by summing the values from the input data patch that has the maximum value. Max-pooling maintains the most significant features of the input by reducing its dimensions.

\begin{equation}
\label{eq3}
    MaxPooling(Y)_{i,j,k} = \max_{m,n}X_{i \cdot s_{x}+ m,j\cdot s_{y}+n,k} 
\end{equation}

In Equation \ref{eq3}, the variable $Y$ represents the input, with $(i,j)$ being the coordinates for the output, and $k$ denoting the channel index. The stride values along the horizontal and vertical axes are represented by $s_x$ and $s_y$ respectively. Additionally, the size of the pooling window is determined by the filter dimensions $f_x$ and $f_y$, which are centered around the output indices $(i,j)$.

Our optimal proposed hybrid model begins with 3 involution layers. Each of them have 3 channels, 3 kernel size and a reduction ratio of 2. These 3 makes the involution block of the proposed model. Then comes the convolution blocks. Each convolution block contains a two-dimensional convolution layer and a two-dimensional max pooling. There are 3 convolution blocks in this model and the first block's convolution layer has 32 nodes. Rest of the block's convolution layers have 64 and 128 nodes respectively. All of them have a kernel size of 3. Each max pooling has a pool size of $2 \times 2$. The max pooling in convolution blocks make the inputs for the Dense layer smaller, as a result our model gets smaller. If we exclude convolutions from our model, the model will become large when we flatten two-dimensional data to one-dimension. But adding convolutions, our model got smaller and we also get the power of convolution in our model. After these 3 convolution blocks, we flatten the information to one dimension. The model has one Dense layer with 128 nodes. There is only one dropout layer in our model which has 10\% dropout rate. The activation function in both involution and convolution block was ReLU. This ends our designing of the proposed model. A simplified visualization of the model can be seen in Figure \ref{model}. This proposed model with 3 involution layers and 3 convolution layers have 356,624 weight parameters and 1.36 MB of storage size. The layer-wise breakdown of output shapes and weight parameters is displayed in Table \ref{tab_1}.

For ablation studies, we only modify the involution block and changing the number of convolution layers drastically change the model size and performance. So, to keep the model small and effective, we only tweak involutions. Involutions specialize eye-tracking data and convolutions easily extract information from the data processed after involution. The effect of involution will be visible in ablation studies.

\begin{figure*}
\centering
\begin{subfigure}{.5\textwidth}
  \centering
  \includegraphics [width=8.7cm, height=6cm]
{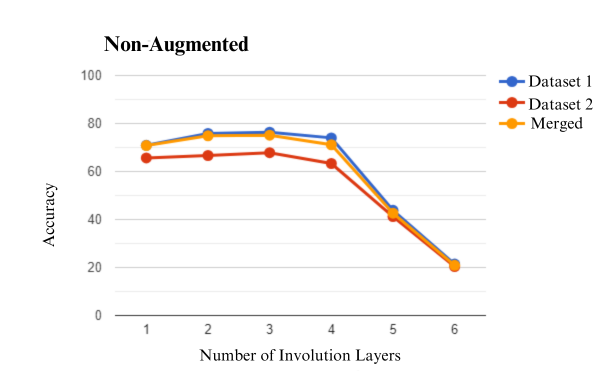}
  \label{fig_7-1}
\end{subfigure}%
\begin{subfigure}{.5\textwidth}
  \centering
  \includegraphics [width=8.7cm, height=6cm]
{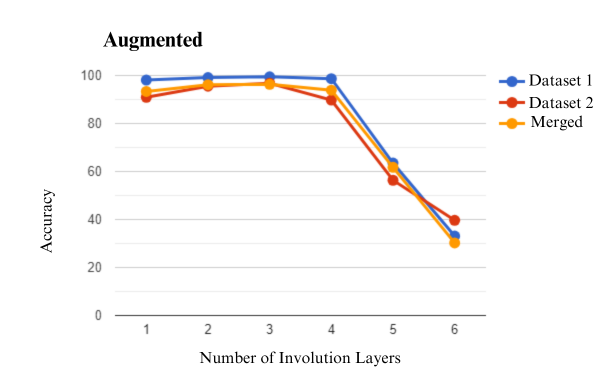}
  \label{fig_7-2}
\end{subfigure}
\caption{Effect of number of involution layers in performance.}
\label{fig_7}
\end{figure*}

\section{Experiments}
\label{Experiments}

In this section, we put the limelight on the setup that we use to experiment, and explain the configuration of the model architectures that are used.

\subsection{Experimental Setup}

We use a few libraries to do all types of training and testing, which include Tensorflow, Keras, and NumPy. Our models undergo performance assessment on various devices, including one equipped with an NVIDIA RTX 2070 offering 7.5 TeraFLOPs of capability, and another featuring an NVIDIA RTX 3080TI GPU, which provides 34.1 TeraFLOPs of performance.

\subsection{Model Specifications}

We use eight different variations of classification models for our ablation study. First, the convolution variant with 3 layers of convolution same as our convolution blocks of the proposed model. Then, we have an involution variant with 3 involution layers same as the proposed model's involution block. Then we have hybrid variations, which has six different models. In this case we increase the number of involution layers by one. In the hybrid models, just like the proposed model, the involution block comes first, then comes the convolution block. The information is flattened to one dimension and a dense layer with 128 nodes proceeds to extract features. Each involution layer takes only 26 weight parameters which is negligible compared to the total size since convolutions use a significant number of weight parameters. We use the same learning rate, number of epochs, and optimizer for all. We use an Adam optimizer with a 0.00001 learning rate and each model is trained with 30 epochs. This number of epochs is optimal since it gives the best fit and performance.

\begin{figure*}
\centering
\begin{subfigure}{.5\textwidth}
  \centering
  \includegraphics [scale=0.6] {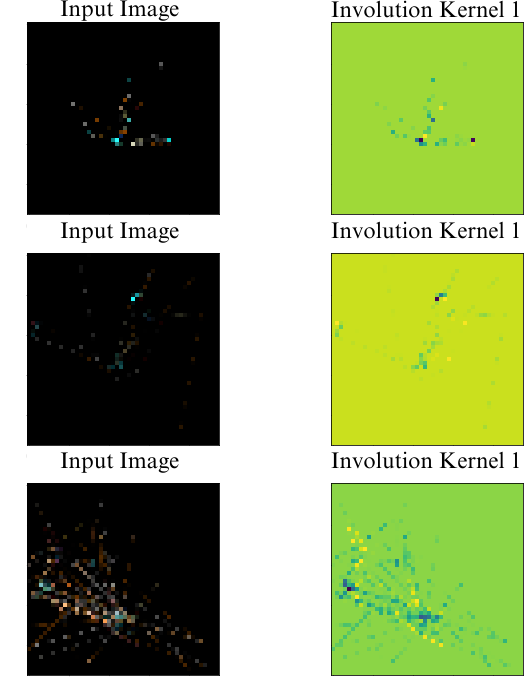}
  \caption{}
\end{subfigure}%
\begin{subfigure}{.5\textwidth}
  \centering
  \includegraphics [scale=0.6] {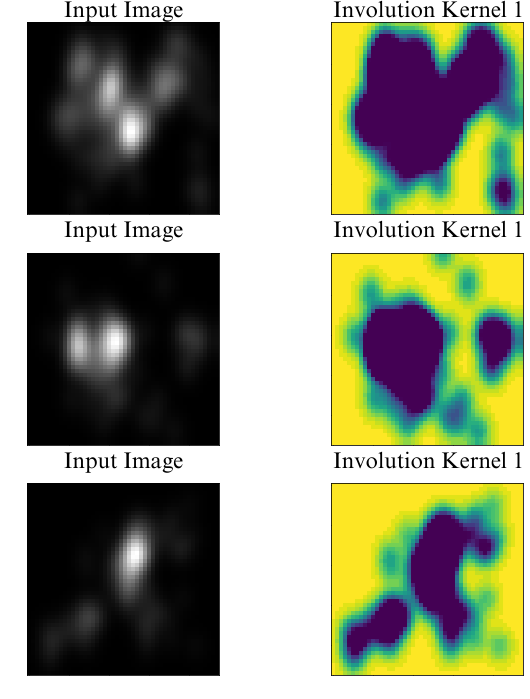}
  \caption{}
\end{subfigure}
\caption{Involution kernel visualizations of hybrid model with 1 involution layer.}
\label{1 layer}
\end{figure*}

\begin{figure*}
\centering
\begin{subfigure}{.5\textwidth}
  \centering
  \includegraphics[scale=0.55] {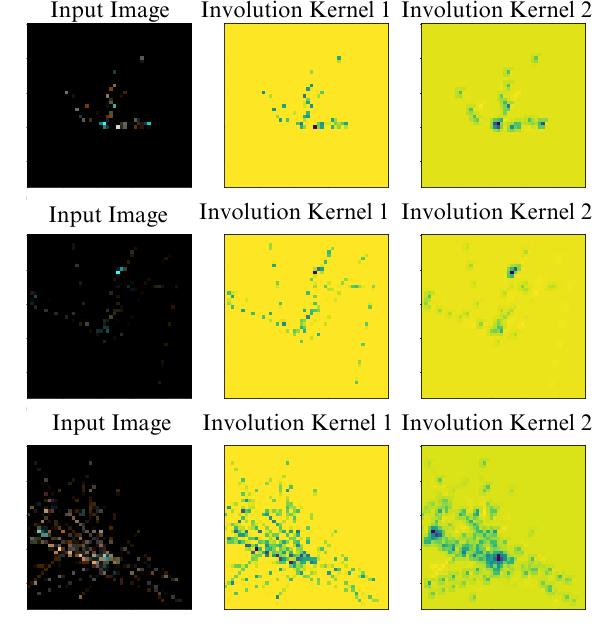}
  \caption{}
  \label{2}
\end{subfigure}%
\begin{subfigure}{.5\textwidth}
  \centering
  \includegraphics[scale=0.55] {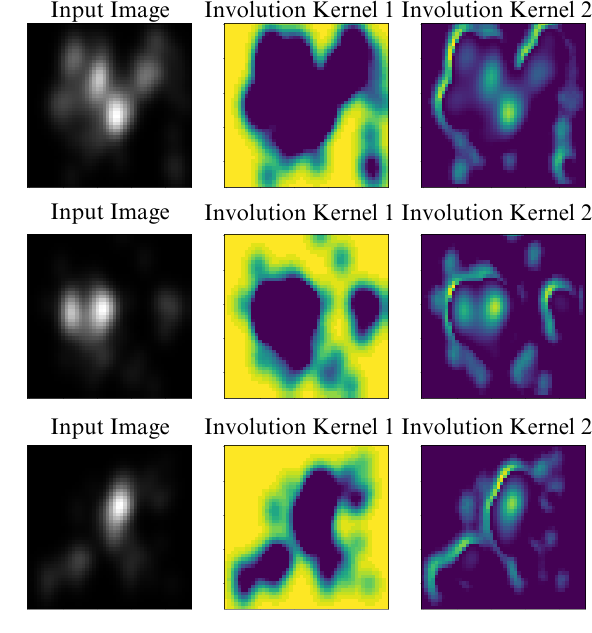}
  \caption{}
  \label{2d}
\end{subfigure}
\caption{Involution kernel visualizations of hybrid model with 2 involution layers.}
\label{2 layer}
\end{figure*}

\begin{figure*}
\centering
\begin{subfigure}{.5\textwidth}
  \centering
  \includegraphics[scale=0.55] {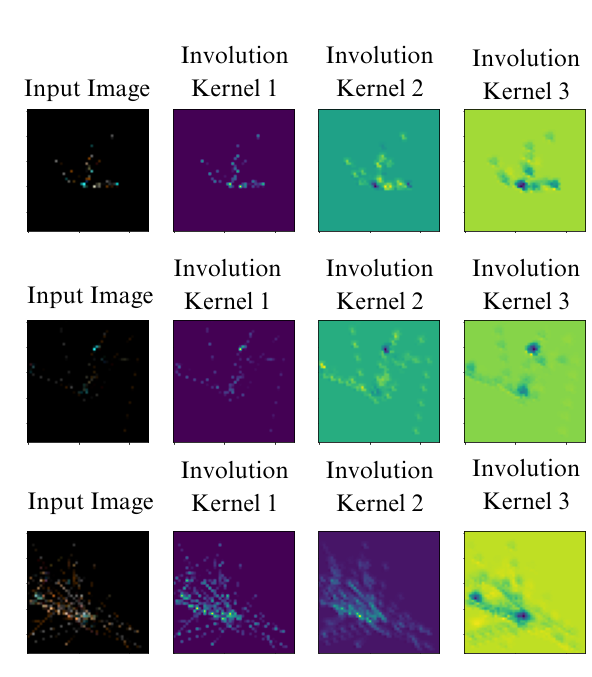}
  \caption{}
  \label{3}
\end{subfigure}%
\begin{subfigure}{.5\textwidth}
  \centering
  \includegraphics[scale=0.55] {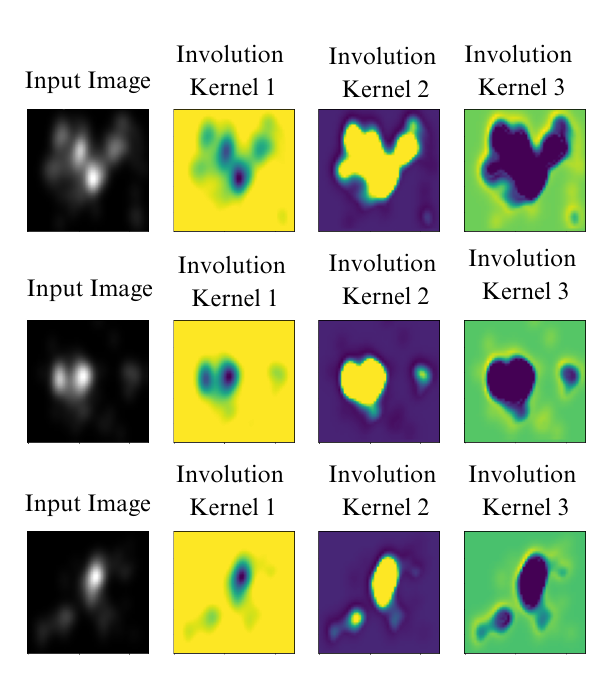}
  \caption{}
  \label{3d}
\end{subfigure}
\caption{Involution kernel visualizations of hybrid model with 3 involution layers.}
\label{3 layer}
\end{figure*}

\begin{figure*}
\centering
\includegraphics [width=\linewidth]{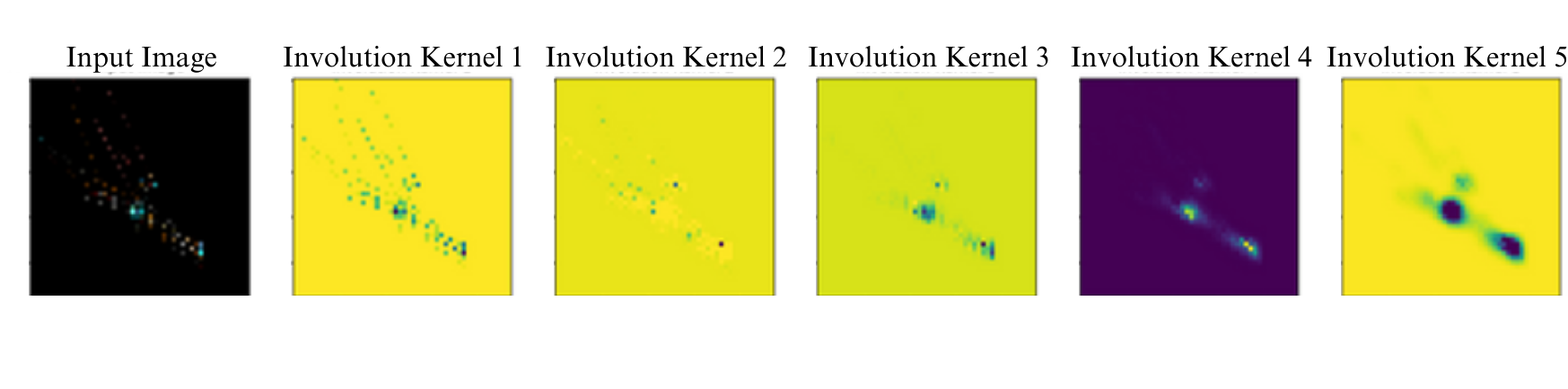}
\caption{Involution kernel visualizations of hybrid model with 5 involution layers.}
\label{5 layer}
\end{figure*}

\begin{figure*}
\centering
\includegraphics [width=\linewidth]{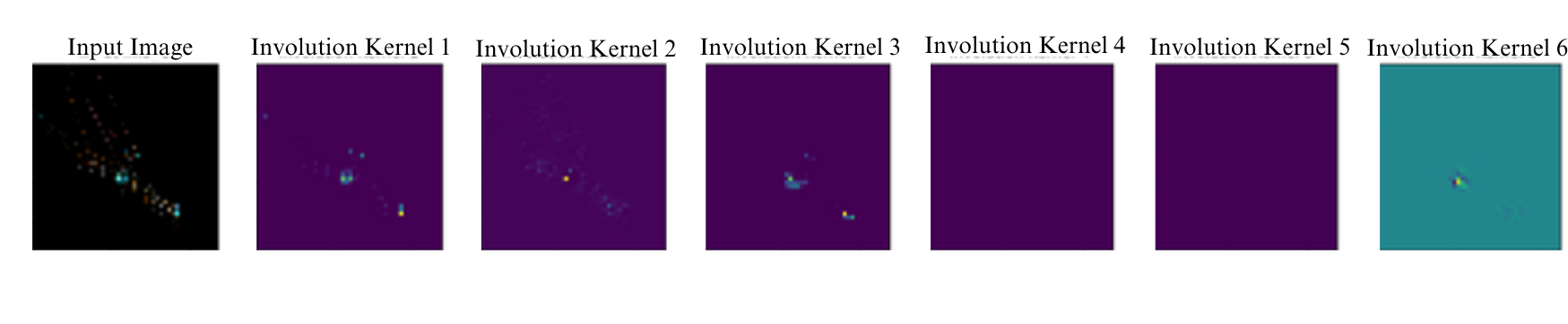}
\caption{Involution kernel visualizations of hybrid model with 6 involution layers.}
\label{6 layer}
\end{figure*}

\subsection{Results and Ablation Studies}

In this section, we will be discussing the performance of the experimented models. We use Test Accuracy, Recall, and F1-score to evaluate the performance of the utilized models.

Table \ref{perform_1} shows the results of non-augmented datasets. The performance of convolution is not the best out of all in any cases. Model with only involutions perform worse than pure convolutions. But the highest accuracy, recall, and F1 score is achieved from the involution-convolution hybrid model with three involution layers. The performance of the hybrid model keeps increasing when we increase the number of involution layers. It indicates, how only one or two involution layers are not enough for extracting the features and learning the exact pattern. Since the location-specific features in the image are distributed throughout the image instead of being at one or few particular regions. Table \ref{perform_2} shows a similar pattern in results. If we look at the convolution-based and involution-based models, they get around 4\% to 5\% less accuracy, recall and F1 score. Moreover, the hybird models tend to achieve high recall rates than their accuracies, which is a good indication since, higher recall rate means lower risks. The performance starts to drop from adding 4 layers, and drastically changes from 5 layers. 

Overall, the performance of hybrid models are effective upto 4 layers, while 3 layers being the optimal amount. In this particular case, more than three layers of involution eventually leads to overfitting and decreases the performance.

Figure \ref{1 layer} shows the kernel visualization of the hybrid model with a single involution layer, which shows that the model is not totally capable to capture the whole pattern from the images. As we increase the involution layers, the location-specific features are captured better than before. Figure \ref{2 layer} and Figure \ref{3 layer} show the improvement in capturing the spatial features of the images, which distributes throughout the images, especially in the case of the ASD class. This kernel visualization shows why a single layer of involution in this particular task, is not sufficient enough. Adding both involution and convolution blocks with three layers of each process makes the task of classifying images much easier for the model. This is a visual representation of showing that  three involution layers are more effective in this task.

While we see a significant boost in performance with the increase in involution layers for both non-augmented and augmented data, the performance quickly falters when the number of layers goes past three. The performance further devolves, worsening than the original accuracy with no involution layers when the number of involution layers becomes five. Inspecting the involution kernels for these layers provides a clue as to why it happens.

While the kernel visualizations of involution architectures with one, two, three, and layers for specific test data images seem to preserve the original features of our target shape, for more than four layers, the visualizations start to lose these features, and thus show their inability to find the target class we are looking for.
We show the loss of features for 5 layers in Figure \ref{5 layer} and for 6 layers in Figure \ref{6 layer} since from 5 layers the negative impact of too many layers are visible in the kernel maps. As each layer looks for more location-specific features, it reverses the original goal, which is finding the pattern that only appears in some parts of the image, and instead starts to pick on noises, even rendering the convolution layers inept to find any more data from this debris. Previous works showed that for some other sources of data \cite{unic}, only one involution layer was enough to boost performance, and the addition of subsequent layers, even two, dramatically decreases the accuracy. Our conjecture is that the exact number of involution layers that maximize classification performance depends on the image features of a particular dataset. What features account for this will have to be subject to further analysis.

\begin{table}[]
\centering
\caption{Hyperparameters and other training details.}
\resizebox{\linewidth}{!}{
\begin{tabular}{ccccc}
\hline
Model & Epochs & Batch Size & Image Size & Learning Rate \\ \hline
VGG16 & 39 & 32 & 100 & 0.000005 \\
ResNet50 & 41 & 32 & 100 & 0.000005 \\
Xception & 39 & 32 & 100 & 0.000005 \\
DenseNet201 & 37 & 32 & 100 & 0.000005 \\
ViT & 25 & 64 & 48 & 0.00001 \\
CCT & 30 & 64 & 48 & 0.00001 \\
INN & 30 & 32 & 48 & 0.00001 \\
Ours & 30 & 32 & 48 & 0.00001 \\ \hline
\end{tabular}%
}
\label{tab_2}
\end{table}

\begin{table}
\caption{Comparison with other popular approaches for dataset-1. Here, the bolded numbers represent the best values.}
\resizebox{.99\linewidth}{!}{
\begin{tabular}{c c c c c c }
\hline
\textbf{Model} & Accuracy  & Recall & F1-Score & Parameters \\ \hline
VGG16          & 83         & 84.04  & 82.43    & 14.7 M      \\ 
ResNet50       & 94.25      & 94.12  & 94.12    & 23.64 M     \\ 
Xception       & 82.18      & 82.18  & 82.33    & 20.8 M      \\ 
DenseNet201    & 91.95     & 92.21   & 92.08    & 18.35 M     \\ 
ViT            & 91        & 92.22   & 91.54    & 21.66 M     \\ 
CCT            & 96.89     & 96.80   & 97.03    & 0.4 M      \\ 
INN            & 97.70     & 97.50   & 97.67    & 0.885 M  \\   
K-Means \cite{18}   & 94     & - & -    & -      \\ 
VAEs \cite{19}  & 70     & -  & -    & -     \\ 
Clustering and MLP \cite{7}    & 87        & -   & -    & -     \\
Custom CNN \cite{Mumenin2023-pt}    & 97.41  & 97.60   & 97.83    & 0.985 M     \\ 
\textbf{Ours }    & \textbf{99.43} & \textbf{99.30} & \textbf{99.49}   &  \textbf{0.356 M}  \\ \hline
\end{tabular}
}
\label{comp1}
\end{table}

\begin{table}
\caption{Comparison with other popular approaches for dataset-2. Here, the bolded numbers represent the best values.}
\resizebox{.99\linewidth}{!}{
\begin{tabular}{c c c c c c }
\hline
\textbf{Model} & Accuracy  & Recall & F1-Score & Parameters \\ \hline

VGG16          & 86.43    & 84.23  & 83.67    & 14.7M      \\ 
ResNet50       & 93.55    & 93.34  & 93.34    & 23.64M     \\ 
Xception       & 80.48    & 79.78  & 80.27    & 20.8M      \\ 
DenseNet201    & 91.95    & 91.71  & 90.08    & 18.35M     \\ 
ViT            & 87.48    & 90.48  & 90.14    & 21.66M     \\ 
CCT            & 96.14    & 95.83  & 95.46    & 0.4 M      \\ 
INN            & 88.34    & 89.89  & 89.67    & 0.885 M    \\
\textbf{Ours }    & \textbf{96.78} & \textbf{96.29} & \textbf{96.58}   & \textbf{0.356 M}  \\ \hline
\end{tabular}
}
\label{comp2}
\end{table}

\begin{table*}[]
\caption{Weight parameters and storage size comparison with other popular approaches. Here, the bolded numbers represent the best values.}
\centering
\resizebox{\textwidth}{!}{%
\begin{tabular}{ccccccccc}
\hline
Model & VGG16 & ResNet50 & Xception & DenseNet201 & ViT & CCT & INN & \textbf{Ours} \\ \hline
Weight Parameters & 14,723,906 &	23,653,250 &	20,898,346 &	18,356,546 &	26,185,673 &	408,139 &	885,200 & \textbf{356,624} \\
Size (Storage) & 56.1672 MB &	90.23 MB &	79.7209 MB &	70.0247 MB & 99.89 MB & 1.56 MB & 3.38 MB & \textbf{1.36 MB} \\ \hline
\end{tabular}%
}
\label{param_comp}
\end{table*}

\begin{table}[h!]
\caption{Performance of non-augmented Data. Here, the bolded numbers represent the best values.}
\centering
\resizebox{.99\linewidth}{!}{
\begin{tabular}{ c c c c c }
\hline
\textbf{Dataset}           & \textbf{Model}              & \textbf{Accuracy} & \textbf{Recall} & \textbf{F1} \\ \hline
\multirow{5}{*}{Dataset-1} & Convolutions (3 layers)                 & 74.42             & 72.39           & 73.00       \\   
                           & Involutions (3 layers)                  & 71.33             & 72.24           & 72.24       \\   
                           & Hybrid (1 layer involution) & 70.91             & 71.18           & 71.25       \\   
                           & Hybrid (2 layers involution) & 75.84            & 76.89          & 77.50      \\   
                           & Hybrid (3 layers involution) & \textbf{76.39}             & \textbf{77.10}           & \textbf{77.65}       
                           \\   
                           & Hybrid (4 layers involution) & 74.01             &    73.92        &     73.62
                           \\   
                           & Hybrid (5 layers involution) & 43.84             &      41.00      &        40.50
                           \\   
                           & Hybrid (6 layers involution) & 21.42             &    20.67       &        20.67 
                           \\ \hline
\multirow{5}{*}{Dataset-2} & Convolutions (3 layers)                 & 57.78             & 57.23           & 57.18       \\   
                           & Involutions (3 layers)                  & 55.32             & 55.50           & 55.29       \\   
                           & Hybrid (1 layer involution) & 65.67             & 65.67           & 65.67       \\   
                           & Hybrid (2 layers involution) & 66.67             & 66.34           & 66.34       \\   
                           & Hybrid (3 layers involution) & \textbf{67.84}             & \textbf{69.66}           & \textbf{68.50}
                           \\   
                           & Hybrid (4 layers involution) & 63.33            & 62.74          & 62.66       
                           \\   
                           & Hybrid (5 layers involution) & 41.22             & 40.89           & 39.64       
                           \\   
                           & Hybrid (6 layers involution) & 20.41            & 18.50           & 18.50       
                           \\ \hline
\multirow{5}{*}{Merged}    & Convolutions (3 layers)                 & 69.57             & 70.23           & 70.15       \\   
                           & Involutions (3 layers)                  & 69.84             & 70.67           & 70.67       \\   
                           & Hybrid (1 layer involution) & 70.81             & 70.99           & 70.85       \\   
                           & Hybrid (2 layers involution) & 74.90             & 75.67           & 75.67       \\   
                           & Hybrid (3 layers involution) & \textbf{75.14}             & \textbf{76.89}           & \textbf{76.33}
                           \\   
                           & Hybrid (4 layers involution) & 71.14             & 71.45           & 71.45       
                           \\   
                           & Hybrid (5 layers involution) & 42.63             & 40.23           & 40.12       
                           \\   
                           & Hybrid (6 layers involution) & 20.89             & 17.50          & 17.50       
                           \\ \hline
\end{tabular}
}
\label{perform_1}
\vspace{-0.5cm}
\end{table}

\begin{table}[]
\caption{Performance of augmented data. Here, the bolded numbers represent the best values.}
\centering
\resizebox{.99\linewidth}{!}{
\begin{tabular}{ c c c c c }
\hline
\textbf{Dataset} & \textbf{Model} & \textbf{Accuracy} & \textbf{Recall} & \textbf{F1} \\ \hline
\multirow{5}{*}{Dataset-1} & Convolutions (3 layers) & 94.44 & 94.18 & 94.25 \\  
                           & Involutions (3 layers)  & 97.70 & 97.50 & 97.67 \\  
                           & Hybrid (1 layer Involution) & 98.09 & 98.79 & 99.10\\  
                           & Hybrid (2 layers Involution) & 99.14 & 99.10 & 99.10 \\ 
                            & Hybrid (3 layers Involution) & \textbf{99.43} & \textbf{99.30} & \textbf{99.49} 
                            \\   
                           & Hybrid (4 layers involution) & 98.67             & 98.79           & 98.55       
                           \\   
                           & Hybrid (5 layers involution) & 63.52             & 60.74           & 60.56       
                           \\   
                           & Hybrid (6 layers involution) & 33.09             & 27.63           & 26.22       
                            \\ \hline
\multirow{5}{*}{Dataset-2} & Convolutions (3 layers) & 93.29 & 92.89 & 92.94 \\   
                          & Involutions (3 layers)  & 88.34 & 89.89 & 89.67 \\   
                           & Hybrid (1 layer Involution) & 90.94 & 89.93 & 90.03 \\   
                           & Hybrid (2 layers Involution) & 95.56 & 94.36 & 95.34 \\   
                          & Hybrid (3 layers Involution) & \textbf{96.78} & \textbf{96.29} & \textbf{96.58} 
                          \\   
                           & Hybrid (4 layers involution) & 89.76             & 89.50           & 89.50       
                           \\   
                           & Hybrid (5 layers involution) & 56.36             & 54.63           & 54.84       
                           \\   
                           & Hybrid (6 layers involution) & 29.67             & 23.62           & 23.54       
                          \\ \hline
\multirow{5}{*}{Merged}    & Convolutions (3 layers) & 92.05 & 91.44 & 91.44 \\   
                            & Involutions (3 layers)  & 93.87 & 93.50 & 93.40 \\   
                           & Hybrid (1 layer Involution) & 93.34 & 93.39 & 93.88 \\   
                           & Hybrid (2 layers Involution) & 96.17 & 96.97 & 96.55 \\   
                           & Hybrid (3 layers Involution) & \textbf{96.32} & \textbf{97.63} & \textbf{97.42} 
                           \\   
                           & Hybrid (4 layers involution) & 93.89             & 94.10           & 93.90       
                           \\   
                           & Hybrid (5 layers involution) & 61.74             & 60.33           & 59.66       
                           \\   
                           & Hybrid (6 layers involution) & 30.26             & 28.36           & 27.89       
                           \\ \hline
\end{tabular}
}
\label{perform_2}
\end{table}

\begin{figure*}
\centering
\begin{subfigure}{.5\textwidth}
  \centering
  \includegraphics [width=8.7cm, height=6cm]
{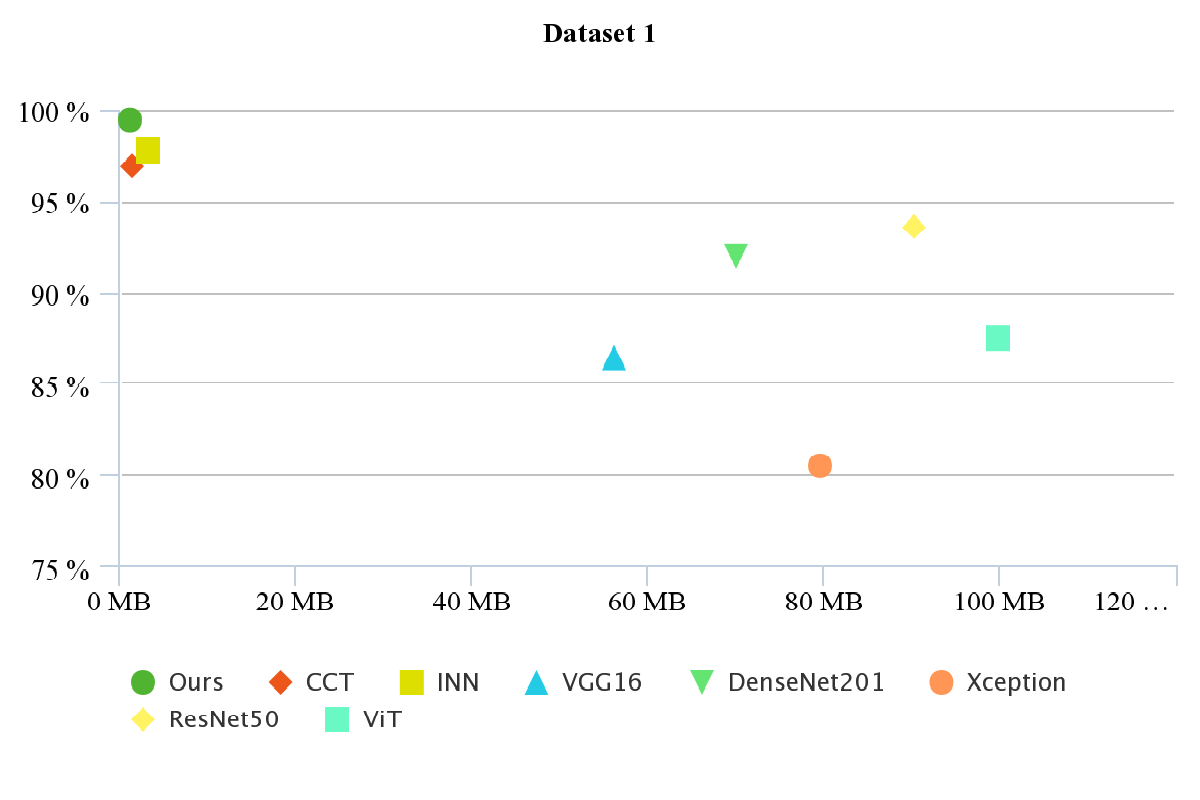}
  \caption{}
  \label{d1}
\end{subfigure}%
\begin{subfigure}{.5\textwidth}
  \centering
  \includegraphics [width=8.7cm, height=6cm]
{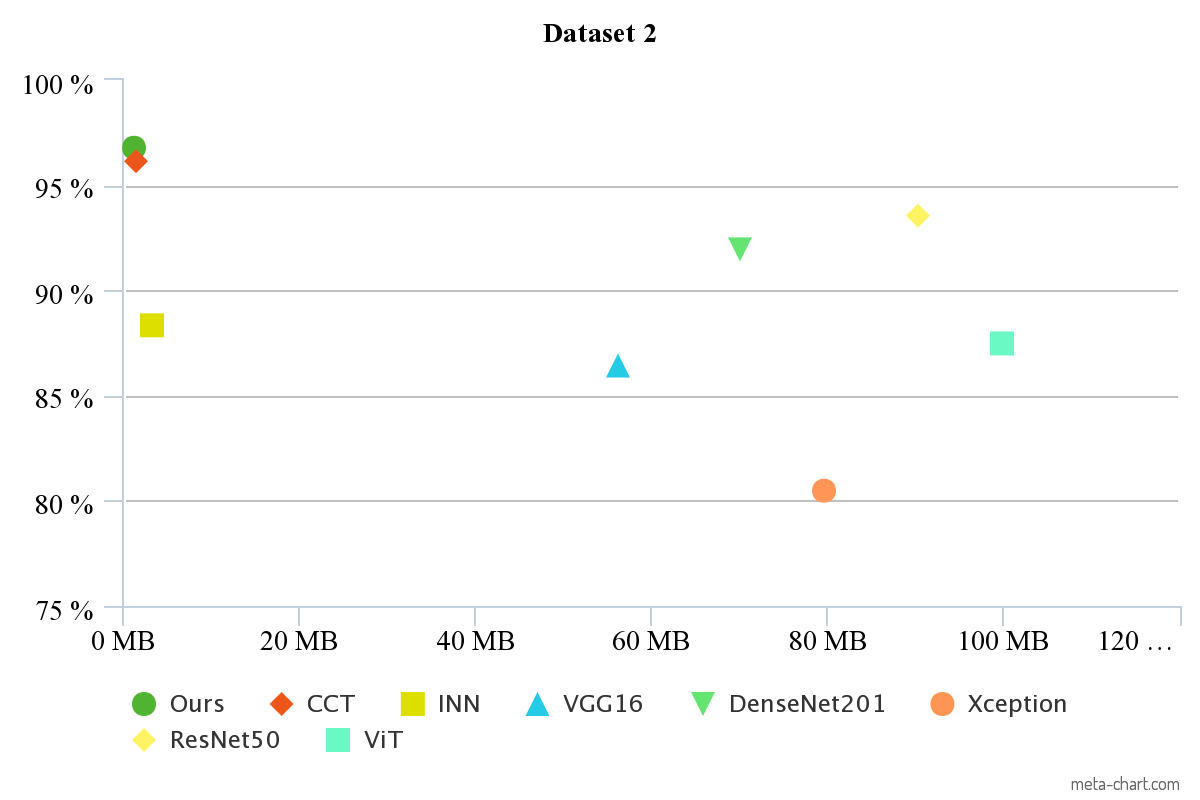}
  \caption{}
  \label{d2}
\end{subfigure}
\caption{Storage size vs accuracy comparison with other models.}
\label{drops}
\end{figure*}

\subsection{Comparison}

For comparison, we utilize Vision Transformer (ViT), Compact Convolutional Transformer (CCT), Involution without the convolution blocks of our proposed model which we refer to as `INN' and popular ConvNet models such as VGG16, ResNet50, Xception, DenseNet201. We use the augmented versions of the datasets in this case, as suggested by the original authors of the dataset and other works. For CCT and ViT, the learning rate is 0.0001, the batch size is 8, and the training was completed in 50 epochs. This is the optimum setup of these transformer-based models for this task. For the ConvNet models, the learning rate is 0.000005, 32 batch sizes, and 35 epochs for completing the training process. Again, this is the best setup for these models. Table \ref{tab_2} represents hyperparameters and training details where VGG16, ResNet50, Xception, DenseNet201 required image size to be 100 pixels to perform well and smaller size is not compatible with these models.

The overall comparison can be seen in Table \ref{comp1} (for dataset-1) and Table \ref{comp2} (for dataset-2). From the tables, we can see that our model outperforms all Even the number of parameters in our proposed model is the lowest. CCT was the second-best in terms of both performance and model parameters. CCT performs almost the same in dataset-2 but the performance in dataset-1 is quite distant. This shows that our model is better in terms of multimodality.

Coming to the model size and requirements, we find that our proposed model has the least storage requirement among all the models. Table \ref{param_comp} shows the weight parameters and storage size comparison with other popular models that we use.

\section{Discussion}
\label{Discussion}

\subsection{Findings Analysis}

We can see that lighter models (CCT, INN and Ours) perform better in this task. The comparison in terms of size can be in Figure \ref{drops}. These models either have location specific operations or attention mechanism, which supports to extract information efficiently. Here, our proposed model outperformed them in both size and performance. The effect of involution layers can be seen in overall quantitative findings from Figure \ref{fig_7} and qualitative findings from Kernel visualizations. The use of convolution with max pooling kept our model compact and efficient since all of the hybrid models upto 4 layers performed better than convolutions with no involution layers. This proves our concept of using involutions for this task, which was mentioned in Data analysis subsection.

Now coming to our model, the reasons involution layers perform best for this data are twofold. First, involution layers are channel-agnostic, and models yield similar performance if the data we are working on here is converted to monochrome, showing a channel-invariance on the data part. This makes a channel-agnostic layer work at least as well as a channel-specific one \cite{10152815}.

Second, involution layers are location-specific within an image. Given the peculiarity of our data, where we are tracking the eye scan-paths and thus the data comes from a very specific set of distribution with bright pixels in very distinct parts of images across the data set, involution layers work the best. Involution layers work better with data distributed exactly in this manner from different domains \cite{involution_paper}.

Since the mean images of the first dataset show that the presence of non-zero pixels is more biased in some locations, we use involution layers exclusively and get a boost in both accuracy and recall. This architecture fail for the second dataset made of saliency maps, as such patterns are not significantly discernible in this dataset. The recall value, however, stay more or less stable, because the data in the ASD class (negative class in this example) are distributed normally, as explained previously, and involution layers have been proven to work on likewise distributed data. Hybrid architectures, in the end, prove to work the best for both datasets, both in terms of accuracy and recall. 

\subsection{Significance}

The signifinace of the proposed architecture are storage size optimization, state-of-the-art performance for eye-tracking or eye scanpath data and its simple implementation. Our study also shows in Figure 8 how lighter models perform better in this task. We also have to note that these lighter models have location specific operations or attention mechanism, which supports to extract information efficiently. Our proposed model is the smallest among others and achieves the best performance. 

The small size of the model may be helful in real world applications. Firstly, implementations on edge devices require smaller models having reduced number of weight parameters to meet their latency requirements \cite{8701, 8702}. In medical image analysis and healthcare informatics, efficiency in DL models are encouraged for smoother real life applications \cite{8703, 8704, 8705}. Moreover, medical solutions with DL requires higher recall to be applicable \cite{8705}, our proposed model has a high recall rate with a reduced number of weight parameters. Finally, mobile applications require smaller optimized models \cite{8706}. ASD diagnosis can be implemented in mobile devices \cite{8707}, also because of new sensor innovations \cite{8708, 8709} in mobile devices the diagnosis of ASD through eye-tracking may begin. Our proposed model is suitable for these applications due to its significantly small size and state-of-the-art performance. 

While there are a number of parameters that contribute to the sustainability and environmental costs associated with AI models, which include hypereparameter optimization \cite{r1} and the performance of the relevant APIs, and while there can be a tradeoff between model size and performance runtime, such that smaller models take up a longer amount of clock cycles to execute a task \cite{3381831}, it is generally agreed that if performance metrics are boosted using a smaller model, that model would be more suited toward environmental protection \cite{r3}. In this aspect, our model can be categorized as a Green AI, as it has a significantly fewer number of parameters when compared to a pure CNN-based model. Our model could be improved in running time with dedicated architectural support, and this lack of tool support is often associated with Green AI \cite{r4}.

\subsection{Limitations}
There are two limitations that we could not include in our research. Firstly, classifying sub-types of ASD. Secondly, measuring FLOPs and MACCs as a metric of computational complexity. Involution and Transformer based architectures are very new comparing to CNN-based models, the libraries are not optimized for calculating FLOPs and MACCs for these new architectures. This limited our contribution in showing the model complexity. But, weight parameters is a significant factor for model complexity, in that way we can say storage-wize our model is the least resource hungry and achieves the best performance comparing to the popular CNN-based and Transformer based models.

\section{Conclusion and Future Work}
\label{Conclusion and Future Work}

The scope of our work is extremely narrow, focusing on the use of deep learning architecture to detect autism-spectrum disorders. The inclusion of involution layers is proven here to surpass the prior performances of other models. Data collected from the eye-tracking of previously diagnosed patients, share some similarities with point cloud data, which is one of the reasons involution is added here. This also shows that models that worked for one specific domain can be adopted for other purposes given enough motivation for their use. We exhibit that involution-convolution hybrid architecture outperforms other models in a field where the convolution approach still reigns supreme while keeping the model size very small. Our future work includes diagnosing different subtypes of ASD using this architecture or improvising the architecture accordingly.

 
\bibliographystyle{elsarticle-num}
\bibliography{cas-refs}


\end{document}